\ifdefined\pdfsuppressptexinfo
\fi
\documentclass{article}
\usepackage{xcolor}
\usepackage{hyperref}

\usepackage[accepted]{mlsys2025}
\renewcommand{\toptitlebar}{\hrule height0pt \vskip .25in}
\renewcommand{\bottomtitlebar}{\vskip .22in \hrule height0pt \vskip .3in}
\mlsystitlerunning{\method: Large-Model Prefill and Small-Model Decoding for LLM Inference via KV Cache Translation}

\usepackage{latexsym}

\usepackage[T1]{fontenc}

\usepackage[utf8]{inputenc}
\usepackage{lineno}

\usepackage{microtype}

\IfFileExists{inconsolata.sty}{\usepackage{inconsolata}}{}

\usepackage{graphicx}
\usepackage{xspace}
\usepackage{booktabs}

\usepackage{caption}
\usepackage{amsmath}
\usepackage{amsfonts}
\usepackage{subfigure}
\usepackage{amssymb}
\usepackage{array,tabularx}
\usepackage{rotating}

\newcommand{\method}{{BridgeKV}\xspace}
\newcommand{\prefillanchor}{\textit{prefill anchor}\xspace}
\newcommand{\decodeanchor}{\textit{decode anchor}\xspace}
\newcommand{\llama}{{Llama-Instruct}}
\newcommand{\llamas}{{Llama-3.2-Instruct}}
\newcommand{\llamam}{{Llama-3.1-Instruct}}
\newcommand{\llamal}{{Llama-2}}
\newcommand{\qwen}{{Qwen2.5-Instruct}}

\begin{document}

\twocolumn[
\mlsystitle{\method: Large-Model Prefill and Small-Model Decoding for LLM Inference via KV Cache Translation}

\mlsyssetsymbol{equal}{*}

\begin{mlsysauthorlist}
\mlsysauthor{Payman Behnam}{equal,to}
\mlsysauthor{Debopam Sanyal}{equal,to}
\mlsysauthor{Pranav Ponnusamy}{to}
\mlsysauthor{Kusum Meda Ravi}{to}
\mlsysauthor{Pranav Sastry}{to}
\mlsysauthor{Mani Chandana Bandaru}{to}
\mlsysauthor{Alind Khare}{goo}
\mlsysauthor{Myungjin Lee}{ed}
\mlsysauthor{Alexey Tumanov}{to}
\end{mlsysauthorlist}

\mlsysaffiliation{to}{Georgia Institute of Technology, USA}
\mlsysaffiliation{goo}{Microsoft; work done while at Georgia Institute of Technology.}
\mlsysaffiliation{ed}{Microsoft; work done while he was employed by Cisco Systems}

\mlsyscorrespondingauthor{Payman Behnam}{payman.behnam@gatech.edu}
\mlsyscorrespondingauthor{Debopam Sanyal}{debopam.sanyal@gatech.edu}
\mlsyscorrespondingauthor{Alexey Tumanov}{atumanov@gatech.edu}

\mlsyskeywords{LLM inference, \method, Multi-Model}

\vskip 0.3in
\begin{abstract}
LLM inference has two phases with different bottlenecks: LLM prefill runs once per request and is compute-intensive. Decoding repeats for every output token and is limited by memory bandwidth.
To address the challenges associated with this issue, different solutions have been proposed. All of these solutions use the same model for both prefill and decode phases.
We propose \method{}, which uses a larger model for prefill and a smaller model of the same family for decoding.
A lightweight KV-stitching module translates the larger model's prompt cache into the smaller model's cache layout.
It handles differences in cache shape and representation through learned local layer mixing and separate nonlinear maps for keys and values.
With both models frozen, we train the module to align their caches and make the smaller model follow the larger model's next-token predictions.
A second training stage adds low-rank adapters when further task adaptation is needed.
Across six model pairs and five datasets, \method{} outperforms fine-tuned small-model baselines with matched trainable parameter budgets in all 30 settings.
It closes up to 80\% of the quality gap to using the larger model for both phases.
On one A40 GPU, \llamam-8B$\rightarrow$\llamas-1B achieves 3.0$\times$ faster requests than the larger model alone, with latency close to the smaller model alone. \method can improve the hardware and system solutions that separate prefill and decode phases across GPUs.    
A separate serving simulation models request arrivals, queues, and cache transfer.
For \llamam-70B$\rightarrow$\llamas-1B, it predicts up to 33.86$\times$ faster requests and 27.5$\times$ lower energy use than using the larger model for both prefill and decode phases.
\end{abstract}]
\printAffiliationsAndNotice{\mlsysEqualContribution}
\section{Introduction}
\label{sec:intro}
In Transformer-based LLM inference, each request runs in two phases that stress hardware differently.
Prefill processes all prompt tokens in parallel and writes their keys and values to the key-value (KV) cache; it is compute-intensive and runs once per request.
Decode then generates one token per step, and each step rereads the model weights and the growing KV cache, so decode is memory-bandwidth-intensive and runs once per output token~\citep{splitwise2024,distserve2024}.
The two phases map directly to the user-facing metrics of LLM serving: prefill largely determines the time to first token (TTFT), and decode determines the time between tokens (TBT).
Because its cost repeats for every output token, decode dominates request latency: for \llamam-8B on one A40 GPU, prefilling a 1K-token prompt takes 0.19~s, while decoding 256 output tokens takes 8.6~s.

This phase asymmetry has motivated prefill-decode disaggregation.
Splitwise~\citep{splitwise2024}, DistServe~\citep{distserve2024}, and Mooncake~\citep{mooncake2025} run prefill and decode on separate GPUs and transfer the KV cache between them, which removes interference between the two phases and raises throughput.
Prefill-as-a-Service~\citep{prefillasaservice2026} goes further and offloads prefill to another datacenter when KV transfer is cheap enough.
All of these systems run the same model on both sides of the transfer.
Multi-LLM serving systems share KV caches across models, but only across models with the same architecture: DroidSpeak~\citep{droidspeak2026}, ICaRus~\citep{icarus2026}, and PrefillShare~\citep{prefillshare2026} reuse caches between a base model and its task-specific fine-tuned variants, which have the same layer count and KV layout.
None of them translates a cache between models with different layer counts or KV dimensions.

Running one model for both phases forces the serving system to choose one point on the latency-accuracy tradeoff. Larger models are more accurate~\citep{kaplan2020scaling,hoffmann2022chinchilla,llama3herd2024}, but every output token must pass through the full large model. Smaller models decode faster and store a smaller KV cache per token, but are less accurate.

\begin{figure}[!t]
\centering
\includegraphics[width=\columnwidth]{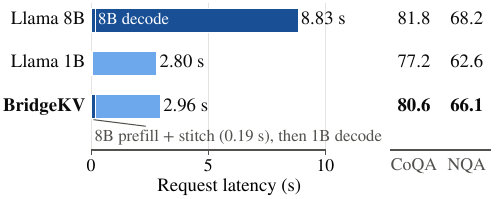}
\vspace{-4ex}
\caption{Request latency and test accuracy of \llamam-8B alone, \llamas-1B alone, and \method with \llamam-8B$\rightarrow$\llamas-1B.
Latency is measured on real hardware (Section~\ref{sec:real_hw}).
The right columns give test F1 on CoQA and NarrativeQA (NQA).}
\label{fig:teaser}
\end{figure}

On CoQA, for example, fine-tuned \llamam-8B reaches 81.8 F1 against 77.2 for fine-tuned \llamas-1B, but its TBT is 4.4$\times$ higher in our serving simulation (Section~\ref{sec:system_perf_model}), and its KV cache takes 4$\times$ more memory per token.
Prefill pays the cost of the model once per request, while decode pays it once per output token.
This asymmetry suggests a phase-aware split: use the larger model for prefill and the smaller model for decode.
The split helps only if much of the accuracy advantage of the larger model is carried in its prompt KV cache.
We ask: \emph{can we use a large model for prefill and a small model for decode, while keeping the cache handoff between both models cheap and the accuracy loss small?}
Figure~\ref{fig:teaser} previews the answer with \method: prefilling with \llamam-8B and decoding with \llamas-1B serves a request in 2.96~s on one A40 GPU, close to the 2.80~s of \llamas-1B alone and 3.0$\times$ faster than \llamam-8B alone, and closes most of the accuracy gap between the two models on CoQA and NarrativeQA.

The main obstacle is the KV cache itself.
A KV cache is model-specific: its number of layers, its KV-head layout, and its representation space match the model that wrote it, so the prompt cache of a larger model is not directly usable by a smaller one.
Even within one model family, the two models store caches of different shapes: \llamam-8B writes 32 layers of keys and values with 128-dimensional heads, while \llamas-1B expects 16 layers with 64-dimensional heads.
The two models also encode the same token in different representation spaces, and both store keys after applying rotary position embeddings (RoPE)~\citep{su2024rope}, so any map on stored keys must also account for position.

Recent work translates KV caches between models so that the receiving model, which takes over generation, can skip its own prefill. Ridge mapper~\citep{ridge2026} fits this translation to reproduce the cache that the receiving model would build from the prompt itself, so it preserves the behavior of the receiving model but does not aim to improve it. In fact, its target is the representation that the receiver would compute itself, without explicit supervision from the larger model's predictions.
Cache-to-Cache~\citep{cachetocache2026} improves the output quality of the receiving model by fusing its cache with the cache of another model, but both models must process the full prompt, so the receiving model still runs its own prefill.
A large-prefill, small-decode split requires two things: the smaller model must skip its own prefill, and it must use the translated cache to mimic the larger model’s next-token predictions instead of making its usual predictions.

\vspace{-2ex}

Lately, model stitching methods connect pretrained models of different sizes through a small learned map between intermediate representations~\citep{snnet2023,esta2024,klas2026, stitchllm2025}.
These works show that the internal representations of related pretrained anchors can be aligned through a small learned map, rather than retraining or distilling an entire new model.

Inspired by these works, we propose to stitch the KV cache in LLM serving.
The larger anchor runs once per request to write the prompt cache, and the smaller anchor reads the stitched cache at every layer and every decode step, with no larger-anchor blocks in the decode loop.
Figure~\ref{fig:overview-method} shows the components and training stages of \method.
The \prefillanchor processes the prompt once and writes its KV cache.
A lightweight KV-stitching module maps this cache into the layout of the \decodeanchor, which then generates every output token on its own after the first token.
For each layer of the \decodeanchor, the module takes a learned weighted combination of a few nearby layers of the \prefillanchor and applies separate maps to keys and values.
Inspired by ShadowKV~\citep{shadowkv}, Empiric~\citep{empiric25}, and Ridge~\citep{ridge2026}, the module removes the RoPE of the \prefillanchor before the key map and applies the RoPE of the \decodeanchor afterward, so the same key map serves every position.
We also follow them in initializing the maps with a closed-form least-squares fit to the cache of the \decodeanchor.
With both anchors frozen, we train the module so that the \decodeanchor, reading the stitched cache, matches the next-token distribution of the \prefillanchor, while a cache-alignment loss keeps the stitched cache close to the native cache of the \decodeanchor.
Low-rank adapters~\citep{lora2022} on the \decodeanchor form a second training stage for tasks that need further adaptation.

This design gives \method three serving advantages.
First, the decode GPU runs the \decodeanchor on a cache with its native shape, just as in a standalone deployment, so TBT matches that of the smaller model.
Second, stitching runs on the prefill GPU, so its negligible overhead affects only TTFT.
Third, stitching occurs before transfer, so the decode GPU receives a cache that incorporates information from the larger model that helps to improve the smaller model’s accuracy.

In summary, \method makes the following contributions:

\begin{figure*}[!ht]
\centering\includegraphics[width=\linewidth]{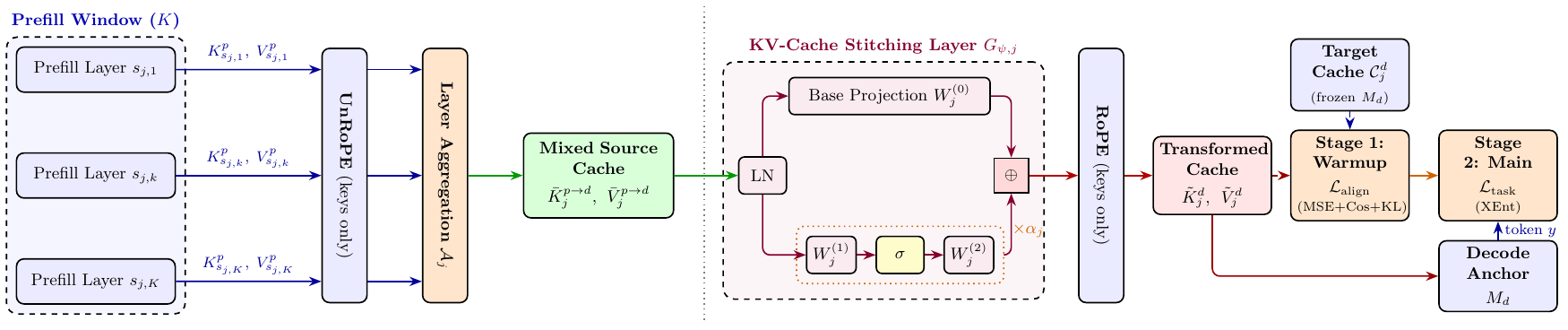}
    \vspace{-5ex}
    \caption{Overview of main components in \method. The left side
    shows windowed layer mapping while the right side illustrates KV
    cache stitching and the two training stages. For keys, the RoPE of
    $M_p$ is removed before the stitching layer and the RoPE of $M_d$
    is applied after it.}
    \label{fig:overview-method}
    \vspace{-2ex}
\end{figure*}
%\vspace{-1ex}

\begin{itemize}
    \item We develop \method{} for large-model prefill and
    small-model decoding within a single request. A learned cache
    handoff transfers the larger anchor's prompt representation
    into the smaller anchor's native KV-cache layout, so generation
    uses only the smaller model.

    \item We introduce a lightweight KV-stitching module with
    learned local layer aggregation and separate residual nonlinear
    maps for keys and values. The module connects anchors with
    different layer counts, KV-head counts, and head dimensions,
    without changing either model's architecture.

    \item We propose a two-stage training recipe that explicitly
    transfers the larger anchor's prediction behavior through the
    stitched cache. The first stage trains only the stitching module,
    combining cache alignment with next-token distillation while
    keeping both anchors frozen. When further task adaptation is
    needed, the second stage jointly trains the stitching module
    and low-rank adapters on the decode anchor.

    \item We evaluate \method{} on five datasets and six model
    pairs. It outperforms fine-tuned small-model baselines with
    matched trainable parameter budgets in all 30 settings and
    closes up to 80\% of the gap to the larger anchor.
    On one A40 GPU, \llamam-8B$\rightarrow$\llamas-1B achieves
    3.0$\times$ faster requests than \llamam-8B alone, taking
    2.96~s compared with 2.80~s for \llamas-1B alone, for a
    1K-token prompt and 256 output tokens.
    Serving simulations further show that \method{} complements
    prefill-decode disaggregation, with up to 33.86$\times$ faster
    requests and 27.5$\times$ lower energy use for
    \llamam-70B$\rightarrow$\llamas-1B compared with the larger
    anchor alone for both phases.
\end{itemize}

 \section{Motivation and Related Work}
 \label{sec:motivrel}

\subsection{Why multi-model prefill and decode?}
\label{sec:whymulti}

For a request with $N_{\mathrm{out}}$ generated tokens, we use the standard decomposition
\vspace{-1ex}
\begin{equation}
T_{\mathrm{response}} = T_{\mathrm{TTFT}} + N_{\mathrm{out}} \cdot T_{\mathrm{TBT}},
\label{eq:response_time}
\vspace{-1.5ex}
\end{equation}
where $T_{\mathrm{TTFT}}$ captures the time to produce the first token and $T_{\mathrm{TBT}}$ is the average time between generated tokens. This decomposition matters because prefill mainly affects $T_{\mathrm{TTFT}}$, while decode mainly affects $T_{\mathrm{TBT}}$ and output throughput.

Different LLMs expose different TTFT and TBT tradeoffs. A larger model often provides stronger prompt understanding and higher task quality, but it also has a higher decode cost because every generated token requires another forward pass through the larger network and transfers of a large KV cache between memory and compute units. A smaller model usually has lower TBT, so it can generate more output tokens per unit time and serve more requests under the same latency deadline. Therefore, using one fixed model for both phases can be inefficient. 
\vspace{-2ex}
\begin{table}[!ht]
\centering
\vspace{0pt}
\small
\caption{Deadline-driven serving motivation under a 2-minute deadline.}
\vspace{-2ex}
\setlength{\tabcolsep}{5pt}
\resizebox{\linewidth}{!}{%
\begin{tabular}{lcccc}
\toprule
Serving strategy & Prefill model & Decode model & Res. time & Served req. \\
\midrule
Large model & \llamam-8B & \llamam-8B & 39.80 s & 3 \\
Small model & \llamas-1B & \llamas-1B &  9.05 s & 13 \\
%Multi-model split & \llamam-8B & \llamas-1B &  9.13 s & 13 \\
%\midrule
%Gain over large-only & -- & -- & -- & $+333.3\%$ \\
\bottomrule
\end{tabular}
}
\vspace{-2ex}
\label{tab:whymulti}
\end{table}

Table~\ref{tab:whymulti} illustrates the latency-quality tradeoff that motivates multi-model serving. 
Using \llamam-8B for the whole request provides the stronger model throughout, but completes only 3 requests within a 2-minute deadline. 
Using \llamas-1B for the whole request completes 13 requests, but processes both the prompt and the generation with the smaller model. 
This suggests a better split: use \llamam-8B for prefill and \llamas-1B for decode. 
This multi-model setting also completes 13 requests, giving 333.3\% more completed requests than large-only serving while preserving large-model prompt processing.

The key challenge is that this split is not possible by simply copying the KV cache from the large model to the small model. The cache produced by the \prefillanchor lives in the representation space of that anchor, while the \decodeanchor expects a cache with its own layer structure and attention geometry. \method addresses this missing piece by learning a lightweight KV cache handoff.

\vspace{-1ex}

\subsection{Related work}
\label{sec:related}
\vspace{-1ex}

\textbf{Prefill-decode disaggregation.}
Splitwise~\citep{splitwise2024}, DistServe~\citep{distserve2024}, and \mbox{Mooncake~\citep{mooncake2025}} place prefill and decode on different resources to match their compute and memory bottlenecks.
Prefill-as-a-Service~\citep{prefillasaservice2026} extends this approach to remote prefill clusters when cache transfer is cheap enough.
These systems separate the phases but keep the same model.
\method instead uses anchors of different sizes, so the handoff must translate the prompt cache into the representation space of the \decodeanchor.

\textbf{Cross-model KV cache reuse.}
DroidSpeak~\citep{droidspeak2026} reuses caches among fine-tuned variants of one base model and recomputes a few sensitive layers.
ICaRus~\citep{icarus2026} and PrefillShare~\citep{prefillshare2026} share a frozen cache-producing prefill module across task-specific decoders, and PrefillShare integrates this sharing into a disaggregated serving system.
aLoRA~\citep{alora2025} lets an activated adapter reuse the prefix cache of its base model.
All of them rely on a common backbone, whereas \method connects distinct anchors with different depths and KV dimensions.
Cache-to-Cache~\citep{cachetocache2026} supports distinct backbones, but fuses their caches after the receiving model also processes the full prompt.

Cache translation is more closely related.
Closed-form ridge maps~\citep{ridge2026} reuse caches between models with matched KV heads and head dimensions, and HeteroFold~\citep{heterofold2026} adds structural alignment and output-aware calibration for cross-family transfer.
RaReCache~\citep{rarecache2026} combines calibrated maps with selective recomputation in the receiving model and also reports large-to-small transfer without recomputation.
Large-prefill, small-decode execution alone therefore does not distinguish \method.
The difference is the training target: these translators fit the cache or the predictions of the receiving model, whereas \method trains the stitched cache so that the smaller model predicts like the larger one.
\method also learns its layer mixture jointly with residual nonlinear maps instead of fixing the source layers in advance.
Appendix~\ref{app:related_comparison} compares these designs and other cache-transfer methods in more detail.

\textbf{Model stitching.}
SN-Net~\citep{snnet2023}, ESTA~\citep{esta2024}, and KLAS~\citep{klas2026} connect pretrained model blocks through lightweight transformations to balance accuracy and cost.
StitchLLM~\citep{stitchllm2025} brings block stitching to LLM serving by combining a bottom block from one model with a top block from another, so its generation path still executes large-model blocks.
\method stitches the prompt cache at the phase boundary, after which decoding uses only the smaller anchor.

\textbf{Small-model adaptation and distillation.}
Supervised fine-tuning~\citep{sft2022}, adapters~\citep{adapter2019}, and LoRA-style methods~\citep{lora2022,qlora2023,dora2024} improve a model through offline adaptation, and distillation methods such as MiniLLM~\citep{minillm2024} transfer the behavior of a teacher into a student during training.
In their standard serving recipes, the adapted small model still processes each prompt itself, whereas \method supplies a stitched cache from the larger model for each request.
The approaches are complementary: our small-anchor baselines are already fine-tuned on each task, \method can optionally add adapters to the \decodeanchor, and its distillation loss trains the KV-stitching module rather than the backbone of the student.

\textbf{Routing, cascades, and speculative decoding.}
RouteLLM~\citep{routellm2025} selects a model for each request, and FrugalGPT~\citep{frugalgpt2024} can escalate a request through a cascade of model calls.
These methods choose which model answers and do not transfer the prompt cache of a large model to a smaller decoder.
Speculative decoding~\citep{leviathan2023speculative,chen2023speculative} uses a small draft model and verification by the large model to preserve the output distribution of the large model, so the large model stays active during generation and the speedup depends on draft acceptance.
\method targets a different tradeoff: after the cache handoff, generation runs only in the small model, without a guarantee of matching the distribution of the large model.
\vspace{-1ex}
\section{Proposed Method}
\label{sec:method}
\subsection{Overview}
\label{sec:overview}

Let $M_p$ be the \prefillanchor and $M_d$ the \decodeanchor, two models of one family that share a tokenizer and have $L_p$ and $L_d$ layers.
Running $M_p$ on a prompt $x_{1:S}$ produces the prompt cache $\mathcal{C}^{p}=\{(K^{p}_{\ell},V^{p}_{\ell})\}_{\ell=1}^{L_p}$ and the first output token.
Running $M_d$ on the same prompt would produce its native cache $\mathcal{C}^{d}=\{(K^{d}_{j},V^{d}_{j})\}_{j=1}^{L_d}$, which differs from $\mathcal{C}^{p}$ in the number of layers, the KV-head layout, and the representation space (Table~\ref{tab:anchors}).
\method replaces the prefill of $M_d$ with a KV-stitching module $G_\psi$ with trainable parameters $\psi$, which maps $\mathcal{C}^{p}$ to a stitched cache $\tilde{\mathcal{C}}^{d}=G_\psi(\mathcal{C}^{p})$ with the shape of $\mathcal{C}^{d}$.
$M_d$ then generates every later token from $\tilde{\mathcal{C}}^{d}$ and appends its own keys and values for each new token, so the cache is stitched only once per request.
In all our experiments, $M_p$ is the larger model, so the larger model runs once per request and the smaller model runs once per output token.

We train $G_\psi$ so that $M_d$, reading $\tilde{\mathcal{C}}^{d}$, predicts like $M_p$ (Section~\ref{sec:training_objective}).
For each layer $j$ of $M_d$, $G_\psi$ first applies a windowed layer mapping that combines a few layers of $M_p$ (Section~\ref{sec:weighted_layer_mapping}) and then key and value stitching layers that map the result into the space of $M_d$ (Section~\ref{sec:kv_layer}):
\begin{equation}
\tilde{K}^{d}_{j}=G^{K}_{\psi,j}\big(\bar{K}^{p\rightarrow d}_{j}\big),
\qquad
\tilde{V}^{d}_{j}=G^{V}_{\psi,j}\big(\bar{V}^{p\rightarrow d}_{j}\big),
\label{eq:kv_transform}
\end{equation}
where $\bar{K}^{p\rightarrow d}_{j}$ and $\bar{V}^{p\rightarrow d}_{j}$ are the keys and values that the mapping assigns to layer $j$: a learned mixture or a concatenation of a few layers of $M_p$.
Figure~\ref{fig:overview-method} shows both components and the two training stages.
The stitched cache is $\tilde{\mathcal{C}}^{d}=\{(\tilde{K}^{d}_{j},\tilde{V}^{d}_{j})\}_{j=1}^{L_d}$.
Section~\ref{sec:inference_serving} describes how \method serves requests.
\vspace{-1ex}
\subsection{Windowed layer mapping}
\label{sec:weighted_layer_mapping}
\vspace{-1ex}

The simplest choice maps each decode layer to the prefill layer at the same relative depth.
However, the prefill layer whose cache best predicts a given decode layer need not sit at that depth, as we find in Appendix~\ref{sec:app_static_mapping}, where we explore other mapping policies.
Therefore, \method gives each decode layer a small window of nearby prefill layers and learns how to combine them, jointly with the stitching layers.
We combine the window either with a learned weighted sum or by concatenation, as described below.

For each decode layer $j \in \{1,\ldots,L_d\}$, \method first finds the prefill layer at the same relative depth,
\begin{equation}
c_j = 1 + \left\lfloor \frac{(j-1)(L_p-1)}{\max(1,\,L_d-1)} \right\rfloor ,
\label{eq:weighted_center}
\end{equation}
and takes the window of $W$ consecutive prefill layers centered on $c_j$, shifted inward at the first and last layers so that it always contains $W$ valid layers:
\begin{equation}
\begin{aligned}
\mathcal{N}_j &= \{s_j,\ldots,s_j+W-1\},\\
s_j &= \max\!\big(1,\,\min(c_j-\lfloor W/2\rfloor,\;L_p-W+1)\big).
\end{aligned}
\label{eq:weighted_candidate_set}
\end{equation}
For \llamam-8B$\rightarrow$\llamas-1B ($L_p=32$, $L_d=16$) with $W=3$, for example, decode layer 1 reads prefill layers 1--3 and decode layer 16 reads prefill layers 30--32.

In the weighted-sum variant, for each decode layer $j$, \method learns separate mixing logits $\beta^{K}_{j},\beta^{V}_{j}\in\mathbb{R}^{W}$ for keys and values, which are part of $\psi$.
The source keys and values for decode layer $j$ are the weighted sums over the window,
\begin{equation}
\begin{aligned}
\bar{K}^{p\rightarrow d}_{j} &= \textstyle\sum_{r=1}^{W} w^{K}_{j,r}\,K^{p}_{s_j+r-1},\\
\bar{V}^{p\rightarrow d}_{j} &= \textstyle\sum_{r=1}^{W} w^{V}_{j,r}\,V^{p}_{s_j+r-1},
\end{aligned}
\label{eq:weighted_cache_mix}
\end{equation}
where $w^{K}_{j}=\operatorname{softmax}(\beta^{K}_{j})$ and $w^{V}_{j}=\operatorname{softmax}(\beta^{V}_{j})$ are the mixing weights for keys and values, respectively.
We initialize the logit of the middle window position to 1 and the others to 0, so the initial weights favor the middle layer of the window, which is $c_j$ except at the first and last layers (for $W=3$, the initial weights are $0.21$, $0.58$, and $0.21$).
Keys and values share the window but have separate weights, so each can draw on different prefill layers.

The stitching layers of Eq.~\eqref{eq:kv_transform} act on each prompt token separately.
For decode layer $j$, $G^{K}_{\psi,j}$ takes the mixed keys of one token, flattened across KV heads into a vector of size $H^{p}_{\mathrm{kv}}d^{p}_{h}$, and outputs a vector of size $H^{d}_{\mathrm{kv}}d^{d}_{h}$, which is reshaped into the head layout of $M_d$; $G^{V}_{\psi,j}$ does the same for values (Section~\ref{sec:kv_layer}).
The mixing weights are trained jointly with the stitching layers.
The \emph{concatenation} variant replaces the weighted sum of Eq.~\eqref{eq:weighted_cache_mix} by concatenating the $W$ window layers along the feature dimension and lets the stitching layer learn the combination.
Its input is $W$ times wider, which is affordable for the Qwen anchors, whose per-layer KV width $H_{\mathrm{kv}}d_h$ is 128 to 512 (Table~\ref{tab:anchors}); we select the variant on the validation split, which picks the weighted sum for every Llama pair and concatenation for every Qwen pair (Appendix~\ref{sec:app_recipes}).

\vspace{-1ex}
\subsection{KV cache stitching layer}
\label{sec:kv_layer}
\vspace{-1ex}
The key and value stitching layers $G^{K}_{\psi,j}$ and $G^{V}_{\psi,j}$ share one residual form with separate parameters:
\begin{equation}
G_{\psi,j}(z) = P_j\bar{z} + b_j + \alpha_j D_j\,\sigma(U_j\bar{z}),
\qquad \bar{z}=\mathrm{LN}(z),
\label{eq:nonlinear_stitch}
\end{equation}
where $z$ is the input vector of one token, $\mathrm{LN}$ is LayerNorm, and $\sigma$ is GELU.
The linear projection $P_j$ maps the KV width of $M_p$ to that of $M_d$, and the branch $D_j\,\sigma(U_j\bar{z})$, whose hidden width is four times the smaller of the two widths, corrects what a linear map misses.
A purely linear map would be cheaper, but we find that linear cross-model KV maps degrade sharply (Appendix~\ref{sec:app_stitching_family}).
The residual scale $\alpha_j$ is learned and starts at $10^{-3}$, so each stitching layer begins as the linear map, and adds the nonlinear correction during training.

\textbf{Rotary position embeddings.} 
Both anchors store keys after applying their rotary position embedding (RoPE)~\citep{su2024rope}, which rotates each key by angles that depend on its position.
The angles also depend on the head dimension, so they differ between the anchors (e.g., 128 for \llamam-8B and 64 for \llamas-1B).
A fixed map applied to rotated keys would in general have to depend on the position of the token.
To avoid this position dependence, we undo the RoPE of $M_p$ on the source keys, apply the key stitching layer, and then apply the RoPE of $M_d$:
\begin{equation}
\tilde{K}^{d}_{j,t} = R^{d}_{t}\, G^{K}_{\psi,j}\big((R^{p}_{t})^{\top} \bar{K}^{p\rightarrow d}_{j,t}\big),
\label{eq:rope_stitch}
\end{equation}
where $t$ is the prompt position, which is the same in both anchors because they share a tokenizer, and $R^{p}_{t}$ and $R^{d}_{t}$ are the RoPE rotations of the two anchors at position $t$.
One stitching layer thus serves every position.
All layers of $M_p$ apply the same rotation $R^{p}_{t}$, so undoing it on each source layer before the weighted sum of Eq.~\eqref{eq:weighted_cache_mix} is equivalent to undoing it on the sum.
RoPE rotates only queries and keys, so the stored values carry no position-dependent rotation and go to the value stitching layer directly.
This follows prior work that operates on keys before RoPE~\citep{shadowkv,empiric25} and the cross-model cache mapping of \citet{ridge2026}.

\textbf{Initialization.}
We initialize the linear projection of every stitching layer in closed form.
We run both anchors on 512 training examples and sample up to 256 positions from each.
For decode layer $j$, the rows of $X_j$ are the inputs of $G_{\psi,j}$ at these positions, computed with the initial mixing weights, with keys un-rotated, and after LayerNorm.
The rows of $Y_j$ are the keys or values that $M_d$ itself writes at the same positions, with keys also un-rotated.
With $\hat{X}_j=[X_j,\mathbf{1}]$, we set
\vspace{-1ex}
\begin{equation}
\begin{bmatrix} P_j^{\top} \\ b_j^{\top} \end{bmatrix}
= \big(\hat{X}_j^{\top}\hat{X}_j + \lambda I\big)^{-1}\hat{X}_j^{\top} Y_j,
\label{eq:ls_init}
\vspace{-1ex}
\end{equation}

where $\lambda$ is $10^{-3}$ times the mean diagonal entry of $\hat{X}_j^{\top}\hat{X}_j$.
$P_j$ maps the full KV width of a token, so it applies directly when the anchors differ in the number of KV heads or in the head dimension.
The fit starts $G_{\psi,j}$ at a regularized least-squares fit to the native cache of $M_d$, and training (Section~\ref{sec:training_objective}) then moves it toward the predictions of $M_p$.
Appendix~\ref{sec:app_stitching_family} compares other stitching-layer designs.

The trainable parameters $\psi$ of the KV-stitching module are, for every decode layer $j$, the mixing logits $\beta^{K}_{j}$ and $\beta^{V}_{j}$ (weighted-sum variant only) and, for both $G^{K}_{\psi,j}$ and $G^{V}_{\psi,j}$, the parameters $P_j$, $b_j$, $U_j$, $D_j$, $\alpha_j$, and those of the LayerNorm.
Both anchors stay frozen while $\psi$ is trained, so in settings that use only this stage, $M_d$ decodes with exactly the weights of the fine-tuned baseline that matches the number of trainable parameters of the stitched model.
For \llamam-8B$\rightarrow$\llamas-1B, $\psi$ has 118M parameters, 12\% of the non-embedding parameters of \llamas-1B, and the module runs only once per request, at the handoff (Section~\ref{sec:inference_serving}).
When a task needs further adaptation, a second stage also trains low-rank adapters $\eta$ on $M_d$ (Section~\ref{sec:training_objective}).

\vspace{-1ex}
\subsection{Training objective}
\label{sec:training_objective}
\vspace{-1ex}
Training has two stages.
The first stage trains only $\psi$, with both anchors frozen, and combines two losses: an alignment loss that keeps the stitched cache close to the native cache of $M_d$, and a distillation loss that makes $M_d$, reading the stitched cache, match the next-token distribution of $M_p$.
The alignment loss alone would at best recover the accuracy of $M_d$; the distillation loss is what carries the accuracy advantage of $M_p$ into the smaller model.
Prior cross-model cache translators instead fit the cache or the outputs of the receiving model itself~\citep{ridge2026,heterofold2026,rarecache2026} (Section~\ref{sec:related}).
The second stage continues training $\psi$ together with low-rank adapters $\eta$ on $M_d$ for the downstream task, when required.

For a training example $(x,y)$, we run $M_p$ on the prompt $x$ to obtain $\mathcal{C}^{p}$ and the stitched cache $\tilde{\mathcal{C}}^{d}=G_\psi(\mathcal{C}^{p})$, and we run $M_d$ on $x$ to obtain its native cache $\mathcal{C}^{d}$.
The alignment loss compares the two caches layer by layer, for keys and for values:
\vspace{-1ex}
\begin{equation}
\mathcal{L}_{\mathrm{align}}(\psi)=\frac{1}{L_d}\sum_{j=1}^{L_d}\sum_{Z\in\{K,V\}}\ell\big(\tilde{Z}^{d}_{j},Z^{d}_{j}\big),
\label{eq:align_loss}
\end{equation}
where $\ell(a,b)=\mathrm{MSE}(a,b)+\lambda_{\cos}\big(1-\cos(a,b)\big)$, the cosine similarity is averaged over corresponding vectors of the two caches, and $\lambda_{\cos}=0.5$.
The MSE term aligns the scale of the cache entries, and the cosine term matches their direction.

The distillation loss compares the predictions of the two anchors on the reference output $y$.
We feed $y$ with teacher forcing to $M_p$, which reads it after its own cache $\mathcal{C}^{p}$, and to $M_d$, which reads it after the stitched cache $\tilde{\mathcal{C}}^{d}$.
With $p^{p}_{t}=p_{M_p}(\cdot\mid \mathcal{C}^{p}, y_{<t})$ and $p^{d}_{t}=p_{M_d}(\cdot\mid \tilde{\mathcal{C}}^{d}, y_{<t})$, the distillation loss is
\vspace{-1ex}
\begin{equation}
\mathcal{L}_{\mathrm{KL}}(\psi)=\frac{1}{|y|}\sum_{t=1}^{|y|}\mathrm{KL}\big(p^{p}_{t}\,\big\|\,p^{d}_{t}\big),
\label{eq:kl_loss}
\end{equation}
and the first-stage objective is
\vspace{-1ex}
\begin{equation}
\min_{\psi}\;\mathcal{L}_{\mathrm{align}}(\psi)+\lambda_{\mathrm{KL}}\,\mathcal{L}_{\mathrm{KL}}(\psi)+\lambda_{\mathrm{rec}}\,\mathcal{L}_{\mathrm{rec}}(\psi).
\label{eq:objective}
\end{equation}
We use $\lambda_{\mathrm{KL}}=0.1$ by default, and the validation search selects 0.05 or 0.5 in a few settings.
In a few settings, we also add an \textit{optional} reconstruction term $\mathcal{L}_{\mathrm{rec}}$ with $\lambda_{\mathrm{rec}}=1$.
The term trains a linear decoder to recover the prefill layer at the center of each window from the stitched cache.
It keeps the module from discarding information that the native cache of $M_d$ lacks.

In the second stage, we train $\psi$ together with low-rank adapters (LoRA)~\citep{lora2022} $\eta$ on $M_d$, using the teacher-forced cross entropy of $M_d$ on the stitched cache:
\vspace{-1ex}
\begin{equation}
\resizebox{0.85\columnwidth}{!}{$\displaystyle
  \mathcal{L}_{\mathrm{task}}(\psi,\eta)
  = -\sum_{t=1}^{|y|}\log p_{M_d}
  \bigl(y_t \mid G_{\psi}(\mathcal{C}^{p}),\, y_{<t};\, \eta\bigr).
$}
\label{eq:task_loss}
\end{equation}
The target $y$ is either the reference output or the output that $M_p$ generates for $x$ (sequence-level distillation~\citep{kim2016seqkd}), and $\mathcal{L}_{\mathrm{KL}}$ can be added with weight 0.5.

For each model pair and dataset, we decide on the validation set whether to use this stage and with which target (Section~\ref{sec:exp_setting}).

We keep the checkpoint with the lowest validation loss; Algorithm~\ref{alg:training} in Appendix~A gives the full training procedure.
We also explore alternative schemes in Appendix~\ref{sec:appendix_ablation}, discussing their advantages and limitations, presenting results and insights, and outlining future directions. Table~\ref{tab:app_summary} provides a summary.

\vspace{-1ex}
\subsection{Inference and serving}
\label{sec:inference_serving}
\vspace{-1ex}
 
At inference time, the evaluation pipeline loads the trained stitched model, loads adapter checkpoint weights, preprocesses the test split with the same tokenizer and formatting policy used during training, generates outputs with a fixed maximum generation length, and reports task accuracy metrics. 
The details of the algorithm are in Appendix~\ref{sec:app_method_algorithms}. \method is compatible with both colocated and disaggregated serving. 
In a colocated setting, $M_p$, $G_{\psi}$, and $M_d$ can run on the same node, avoiding network transfer and focusing on compute and latency tradeoffs. 
In a disaggregated setting, the transformed cache can be transferred to decode workers, similar to prior prefill-decode systems~\citep{splitwise2024,distserve2024}.  The key difference is that \method transfers a decode-anchor-compatible cache, enabling the decode cluster to run the smaller model.

\vspace{-1ex}
\section{Experiments and Results}
\label{sec:results}

\subsection{Experimental setting}
\label{sec:exp_setting}
We evaluate \method on multi-model prefill-decode pairs from two instruction-tuned model families.
For the Llama family, we use \llamas{} models at 1B and 3B scales and \llamam-8B as the larger anchor~\citep{llama3herd2024,llama32modelcard2024}.
For the Qwen family, we use \qwen{} models at 0.5B, 3B, and 7B scales~\citep{qwen25technical2024}.
We consider six multi-model handoff settings.
For Llama-Instruct, we evaluate 3B$\rightarrow$1B, 8B$\rightarrow$1B, and 8B$\rightarrow$3B handoffs.
For Qwen-Instruct, we evaluate 3B$\rightarrow$0.5B, 7B$\rightarrow$0.5B, and 7B$\rightarrow$3B handoffs.
In each pair, the first model denotes the larger \prefillanchor and the second model denotes the smaller \decodeanchor, and the anchors of one family share a tokenizer.
Table~\ref{tab:anchors} lists the layer count, KV layout, and per-token KV-cache size of each anchor.

\begin{table}[t]
\centering
\small
\setlength{\tabcolsep}{4pt}
\caption{The six anchors. $L$: layers; $H_{\mathrm{kv}}{\times}d_h$: KV heads $\times$ head dimension; $\kappa$: KV-cache bytes per token in bf16.}
\vspace{-2ex}
\begin{tabular}{lrrr}
\toprule
Model & $L$ & $H_{\mathrm{kv}}{\times}d_h$ & $\kappa$ (KiB) \\
\midrule
\llamas-1B & 16 & 8$\times$64  & 32 \\
\llamas-3B & 28 & 8$\times$128 & 112 \\
\llamam-8B & 32 & 8$\times$128 & 128 \\
\midrule
\qwen-0.5B & 24 & 2$\times$64  & 12 \\
\qwen-3B   & 36 & 2$\times$128 & 36 \\
\qwen-7B   & 28 & 4$\times$128 & 56 \\
\bottomrule
\end{tabular}
\vspace{-2ex}
\label{tab:anchors}
\end{table}
%\vspace{-1ex}

We evaluate \method on several datasets from LongBench~\citep{longbench2024}.
We select representative tasks that stress different long-context abilities: single-document question answering, summarization, and few-shot learning.
For single-document QA, we use NarrativeQA~\citep{kocisky2018narrativeqa}, with the plot summary of each story as the context.
For summarization, we use GovReport~\citep{huang2021govreport} and Multi-News~\citep{multinews2019}.
For few-shot learning, we use SAMSum~\citep{gliwa2019samsum}.
We also evaluate on CoQA~\citep{reddy2019coqa}, a conversational QA dataset outside LongBench, where each question also depends on the earlier turns.
Appendix~\ref{sec:app_datasets} gives the prompt lengths, output lengths, and test-set sizes.

We evaluate the system-level behavior of \method using a Splitwise-style discrete-event simulator~\citep{splitwise2024} to model machine pools, queues, schedulers, and KV-cache transfer.
More details are in Section~\ref{sec:system_perf_model} and Appendix~\ref{sec:appendix_sys_results-overview}.\footnote{We release the code upon acceptance of the paper.}

\vspace{-1ex}
\subsubsection{Hyperparameter tuning}
%and training stability}
\vspace{-1ex}
We describe the key hyperparameters governing the windowed layer mapping, stitching layer capacity, and optimization schedule, along with the empirical rationale for each choice.
For each model pair and dataset, we select the configuration from these options on the validation split and evaluate it once on the test split.
Appendix~\ref{sec:app_recipes} lists the training stages selected for every setting.

\textbf{Window size.}
The windowed layer mapping uses a local window of $W=3$ prefill layers per decode layer by default.
A window of $W=1$ reduces to a rigid one-to-one mapping that cannot compensate for representational depth mismatch between anchors.
Larger windows ($W \geq 6$) increase the number of trainable mixing weights without consistent downstream gains, suggesting that the relevant source information is concentrated in a narrow neighborhood.
$W=3$ provides sufficient flexibility to interpolate across the local depth neighborhood while keeping the mixing-weight overhead minimal.

\textbf{Stitching layer capacity.}
To prevent overfitting when mapping between distinct latent spaces, we constrain the LoRA rank of the second stage to $r \in \{8, 16\}$ and fix the scaling factor to $\alpha = 2r$.
The MLP hidden expansion factor $\gamma$ controls the intermediate dimensionality between $U_j$ and $D_j$; a larger $\gamma$ increases the capacity of the nonlinear branch but risks overfitting to the training cache distribution, while a smaller $\gamma$ keeps the transformation closer to the base linear projection.
$\gamma=4$ provides sufficient capacity for the nonlinear correction while keeping the stitching layers small, so we fix it for every model pair.
The LoRA adapters use dropout 0.1, and the stitching layers use no dropout.

\textbf{Residual scale initialization.}
We initialize $\alpha_j=10^{-3}$ for every model pair (Section~\ref{sec:kv_layer}), small enough to avoid large perturbations in the early optimization steps while still letting the nonlinear correction grow during training.

\textbf{Loss weighting.}
The cosine similarity weight $\lambda_{\cos}$ balances numerical and directional alignment between the stitched and native caches, and a single shared value of $\lambda_{\cos}=0.5$ works for both model families.
The weight of the distillation term is selected from $\lambda_{\mathrm{KL}}\in\{0.05, 0.1, 0.5\}$, with $0.1$ as the default, and the reconstruction term is either off or used with $\lambda_{\mathrm{rec}}=1$.

\textbf{Optimization schedule.}
Stable gradient estimates require a minimum batch size of 16; smaller batches produced noisy alignment loss and inconsistent convergence, so we use an effective batch size of 32.
Training proceeds in two stages: a first stage of 1 to 5 epochs with a learning rate of $5\times10^{-4}$, during which only $G_\psi$ receives gradients to align the two representation spaces, followed by an optional second stage of 1 or 2 epochs with a learning rate in $\{5\times10^{-5}, 10^{-4}, 2\times10^{-4}\}$ and weight decay $0.05$.
The first stage is necessary because the optimization landscape for cross-model cache alignment is considerably harder than standard task fine-tuning, and premature transitions to the task loss sharply lowered accuracy in preliminary experiments (Appendix~\ref{sec:app_kd}).

\vspace{-1ex}
\subsection{Task-quality results}
\label{sec:quality_plan} 
\vspace{-1ex}
\begin{figure*}[t]
\centering
\includegraphics[width=0.9\textwidth]{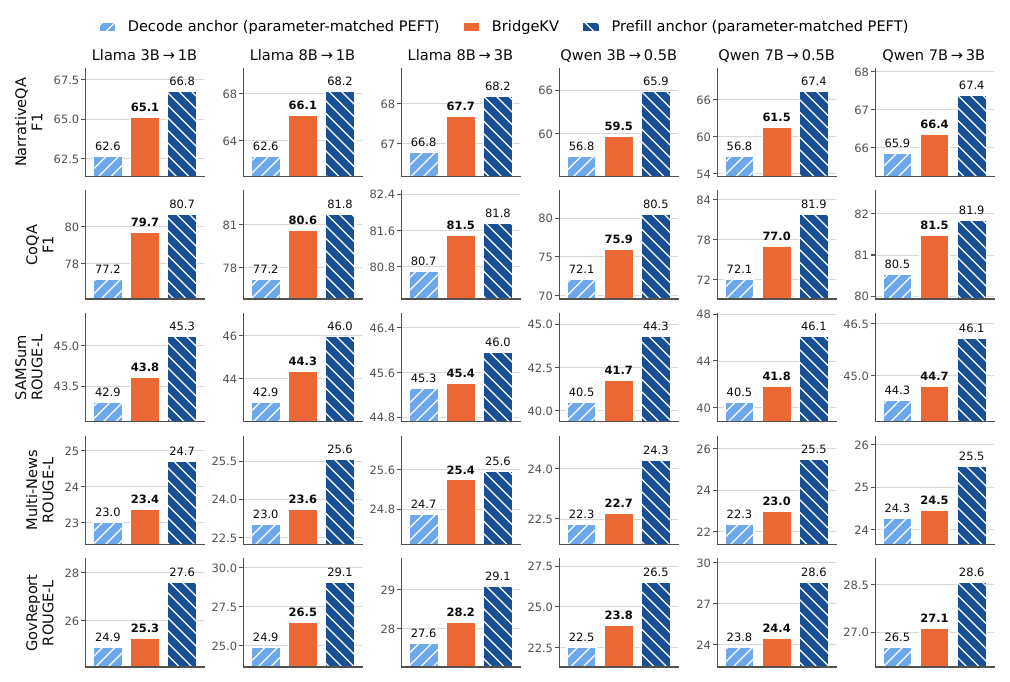}
\vspace{-3ex}
\caption{Test accuracy of the \decodeanchor, \method, and the \prefillanchor for each model pair (columns) and dataset (rows), measured with token F1 on NarrativeQA and CoQA and with ROUGE-L on the other datasets.
Both anchors are fine-tuned with LoRA at a rank that matches the number of trainable parameters of the stitched model, so the gain of \method does not come from training more parameters.
The \decodeanchor generates every output token of \method after the first one.
The y-axes do not start at zero and differ between panels.}
\label{fig:accuracy_stitch}
\vspace{-2ex}
\end{figure*}

We compare \method with its two anchors, each fine-tuned on the task with LoRA, using the same number of trainable parameters as the stitched model.
\method uses these fine-tuned anchors as its \prefillanchor and \decodeanchor.
We report the metrics that LongBench~\citep{longbench2024} uses: token F1 on NarrativeQA and ROUGE-L~\citep{rouge2004} on SAMSum, Multi-News, and GovReport.
CoQA is not part of LongBench, so we report its official metric, token F1~\citep{reddy2019coqa}, computed with the same normalization.
We select configurations and decoding on the validation split with the same metrics.

Figure~\ref{fig:accuracy_stitch} shows how the proposed \method can increase the accuracy of the \decodeanchor across various datasets and models.
For every plotted setting, \method improves over the fine-tuned \decodeanchor while keeping the same small model for generation.
However, the magnitude of the gain depends on the task and model pair.
As shown, \method improves the accuracy of the directly fine-tuned \decodeanchor by up to 8.4\% relative (NarrativeQA, for \qwen-7B$\rightarrow$\qwen-0.5B).
In the worst case, the relative improvement is 0.2\% (SAMSum, for \llamam-8B$\rightarrow$\llamas-3B), while \method can reach almost the same latency, throughput, and energy efficiency as serving with the smaller model (Section~\ref{sec:system_perf_model}).
This is achieved with low retraining costs. In 12 of the 30 settings, the selected configuration uses only the first stage of training (Appendix~\ref{sec:app_recipes}), so $M_d$ decodes with exactly the weights of its fine-tuned baseline, and \method still scores higher in all 12.
In these settings, the gain comes from the stitched cache alone, not from extra trainable parameters on the decode side.

\begin{table}[t]
\centering
\caption{Training cost in A40 GPU-hours of the two stages of the KV-stitching module and of a PEFT baseline that fine-tunes the \decodeanchor with LoRA at a parameter-matched rank. ``--'': the setting uses only the first stage.
The PEFT rank is matched to the stitched model of each row, so it differs between rows with the same \decodeanchor, and training stops early on the validation loss.}
\vspace{-2ex}
\small
\setlength{\tabcolsep}{4pt}
\begin{tabular}{llrrr}
\toprule
Dataset & Pair & Stage 1 & Stage 2 & PEFT \\
\midrule
NarrativeQA & Llama 3B$\rightarrow$1B & 2.1 & -- & 2.4 \\
 & Llama 8B$\rightarrow$1B & 3.3 & -- & 3.4 \\
 & Qwen 3B$\rightarrow$0.5B & 1.7 & 1.7 & 3.9 \\
 & Qwen 7B$\rightarrow$0.5B & 2.7 & 1.1 & 4.3 \\
\midrule
SAMSum & Llama 3B$\rightarrow$1B & 0.4 & 0.4 & 1.2 \\
 & Llama 8B$\rightarrow$1B & 1.6 & 0.6 & 2.4 \\
 & Qwen 3B$\rightarrow$0.5B & 1.0 & 0.4 & 2.4 \\
 & Qwen 7B$\rightarrow$0.5B & 1.5 & 0.6 & 2.4 \\
\midrule
GovReport & Llama 3B$\rightarrow$1B & 8.2 & -- & 37.8 \\
 & Llama 8B$\rightarrow$1B & 12.1 & -- & 37.8 \\
 & Qwen 3B$\rightarrow$0.5B & 6.2 & 5.1 & 31.3 \\
 & Qwen 7B$\rightarrow$0.5B & 9.7 & 2.1 & 24.8 \\
\midrule
Multi-News & Llama 3B$\rightarrow$1B & 5.2 & 7.5 & 23.5 \\
 & Llama 8B$\rightarrow$1B & 7.7 & 4.1 & 22.3 \\
 & Qwen 3B$\rightarrow$0.5B & 6.9 & 3.7 & 14.1 \\
 & Qwen 7B$\rightarrow$0.5B & 6.2 & 2.0 & 15.1 \\
\bottomrule
\end{tabular}
\label{tab:train_cost}
\vspace{-1ex}
\end{table}
\vspace{-1ex}
Table~\ref{tab:train_cost} compares the training cost of the KV-stitching module with that of the parameter-matched PEFT baseline, which fine-tunes the \decodeanchor alone with LoRA at the same number of trainable parameters as the stitched model.
Training the module is cheaper in every setting.
The PEFT baseline therefore receives the same number of trainable parameters and a larger training budget than the KV-stitching module, and \method still scores higher in every setting.

\vspace{-1ex}
\subsection{Experiments on real hardware}
\label{sec:real_hw}

\begin{figure}[t]
\centering
\includegraphics[width=0.85\columnwidth]{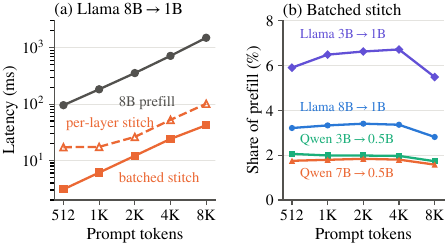}
\vspace{-2ex}
\caption{Cost of the KV-stitching module, measured on real hardware (one A40 GPU, batch size 1).
(a) For \llamam-8B$\rightarrow$\llamas-1B, prefill time of the 8B anchor and stitching time of the per-layer and batched implementations, on a log scale.
(b) Stitching time of the batched implementation as a share of the prefill time of the larger anchor, for the four pairs with a 1B or 0.5B \decodeanchor; all modules use a window of $W=3$.}
\label{fig:handoff}
\vspace{-2ex}
\end{figure}

This subsection measures on real hardware what the KV handoff of \method costs and how long a complete request takes.
We run the anchors and the KV-stitching module in bf16 with Hugging Face Transformers on one NVIDIA A40 GPU at batch size 1, and report the median of 20 runs after 3 warmup runs.
We measure the four pairs with a 1B or 0.5B \decodeanchor, at prompt lengths from 512 to 8K tokens.

A direct implementation of $G_\psi$ follows Section~\ref{sec:kv_layer} layer by layer.
This \emph{per-layer implementation} loops over the $L_d$ decode layers and over keys and values, so it launches $2L_d$ chains of small GPU kernels, and the fixed launch cost of each kernel dominates its running time.
All stitching layers of a pair have the same shapes, so the \emph{batched implementation} stacks their weights and runs the whole module in one pass.
The pass undoes the RoPE of $M_p$ on every source key once, forms the input of every decode layer, the weighted sum of Eq.~\eqref{eq:weighted_cache_mix} or the concatenation of its window (Section~\ref{sec:weighted_layer_mapping}), with one batched operation, applies LayerNorm, evaluates Eq.~\eqref{eq:nonlinear_stitch} for all layers with three batched matrix multiplications, and applies the RoPE of $M_d$; we compile the pass with \texttt{torch.compile}~\citep{ansel2024pytorch2}.

The batched implementation adds 1.6--6.7\% to the prefill time of the \prefillanchor across the four pairs, whereas the per-layer implementation adds 5--51\% (Figure~\ref{fig:handoff}b).
For \llamam-8B$\rightarrow$\llamas-1B, the batched implementation takes 3--42~ms against 96--1504~ms of prefill, and its time grows with prompt length at roughly the same rate as prefill (Figure~\ref{fig:handoff}a).
The module runs on the prefill GPU before the transfer, so this cost adds only to TTFT.
We also measure the end-to-end latency of one \llamam-8B$\rightarrow$\llamas-1B request with a 1K-token prompt and 256 output tokens, where prefill emits the first output token.
\method prefills with \llamam-8B (0.19~s), stitches the cache (6~ms), and decodes with \llamas-1B at 10.9~ms per token, against 33.9~ms per token for \llamam-8B.
The request takes 2.96~s, close to the 2.80~s of \llamas-1B alone and 3.0$\times$ faster than the 8.83~s of \llamam-8B alone (Figure~\ref{fig:teaser}).

\vspace{-1ex}
\subsection{System-side performance model}
\label{sec:system_perf_model}
\vspace{-1ex}
We evaluate the system-level behavior of \method using the same Splitwise-style simulator~\citep{splitwise2024}. The main goal here is to show that our approach can be combined with disaggregated hardware solutions (e.g., Splitwise~\citep{splitwise2024}, DistServe~\citep{distserve2024}) to further improve them.
Its results differ from the measurements of Section~\ref{sec:real_hw} because the simulator models a disaggregated deployment that serves many requests at once and includes queueing delay, KV cache transfer overhead, etc.
Unless otherwise stated, the main system results use a 200-request trace with 1024 input tokens and 1024 output tokens per request. 
Request arrivals follow an exponential distribution with an arrival rate of 2.0 requests/s.
We keep the \llamam-8B$\rightarrow$\llamas-1B setting in the main paper and discuss the results when the parameter gap between \prefillanchor and \decodeanchor increases, and move other models and sensitivity studies to Appendix~\ref{sec:appendix_sys_results-overview}. 
We report response time, TTFT, TBT, throughput, and energy. Throughput is computed from the decode rate.

During prefill, each Transformer layer creates key and value tensors.  
The per-layer KV cache size is
\vspace{-1ex}
\begin{equation}
M_{\mathrm{KV}}^{\mathrm{layer}} =
B_{\mathrm{sz}} \cdot S \cdot H_{\mathrm{KV}} \cdot d \cdot 2 \cdot b,
\label{eq:kv_size_per_layer}
\end{equation}
where $B_{\mathrm{sz}}$ is batch size, $S$ is prompt length, $H_{\mathrm{KV}}$ is the number of KV heads, $d$ is the head dimension, the factor 2 accounts for keys and values, and $b$ is bytes per scalar. 
The total prompt KV cache transferred from the prefill anchor is
\vspace{-1ex}
\begin{equation}
M_{\mathrm{KV}}^{\mathrm{total}} =
L \cdot M_{\mathrm{KV}}^{\mathrm{layer}},
\label{eq:kv_size_total}
\vspace{-1.5ex}
\end{equation}
where $L$ is the number of prefill-anchor layers.

The raw transfer latency is
\vspace{-1ex}
\begin{equation}
T_{\mathrm{KV}} =
\frac{M_{\mathrm{KV}}^{\mathrm{total}}}{B_{\mathrm{link}}},
\label{eq:kv_raw_transfer}
\vspace{-1.5ex}
\end{equation}
where $B_{\mathrm{link}}$ is the effective interconnect bandwidth. 
We use 100 GB/s for NVLink and 50 GB/s for InfiniBand, following the same communication model used in our system-side study. 
Since Splitwise-style execution transfers KV cache layer by layer while later prefill layers are still being computed, the visible overhead is approximated by the final non-overlapped layer transfer:
\vspace{-1ex}
\begin{equation}
T_{\mathrm{visible\ KV}} =
\frac{T_{\mathrm{KV}}}{L}
=
\frac{M_{\mathrm{KV}}^{\mathrm{layer}}}{B_{\mathrm{link}}}.
\label{eq:visible_kv}
\vspace{-1.5ex}
\end{equation}

For the main \llamam-8B$\rightarrow$\llamas-1B setting, we use $B_{\mathrm{sz}}=1$, $S=1024$, $L=32$, $H_{\mathrm{KV}}=8$, $d=128$, and $b=2$. 
Even under InfiniBand, the visible layer-wise overhead is only 0.08 ms.
\vspace{-1ex}
\begin{table}[!ht]
\centering
\caption{KV cache transfer overhead for 1024-token prefill. Layer-wise overlap makes only one layer transfer visible.}
\vspace{-2ex}
\small
\resizebox{\columnwidth}{!}{%
\begin{tabular}{lcccccc}
\toprule
Pair &  Total KV & Raw NVLink & Visible NVL & Raw IB & Visible IB \\
\midrule
Llama-8B$\rightarrow$1B &  0.13 GiB & 1.34 ms & 0.04 ms & 2.68 ms & 0.08 ms \\
\bottomrule
\end{tabular}
}
\vspace{-2ex}
\label{tab:kv_transfer_predicted}
\vspace{-2ex}
\end{table}

Table~\ref{tab:main_8b1b_latency} reports the system result for \llamam-8B$\rightarrow$\llamas-1B with 1024 input and 1024 output tokens over 200 requests. 
The larger \llamam-8B anchor is used for prefill, while the smaller \llamas-1B anchor is used for decode after the learned KV cache handoff. 
Because the workload generates 1024 output tokens, the repeated decode cost dominates response time.
\vspace{-1ex}
\begin{table}[!ht]
\centering\small
\caption{Latency and throughput for \llamam-8B$\rightarrow$\llamas-1B,
1024 input/1024 output.}
\vspace{-2ex}
\resizebox{0.99\columnwidth}{!}{%
\begin{tabular}{lccccc}
\toprule
Model & Response & TTFT & TBT & Throughput & Thr. Imp. \\
\midrule
Llama-1B           &  9.05 s &  27.5 ms &  8.81 ms & 113.5 tok/s & 4.40$\times$ \\
Llama-8B           & 39.80 s & 129.3 ms & 38.74 ms &  25.81 tok/s & baseline \\
8B$\rightarrow$1B    &  9.13 s & 111.4 ms &  8.81 ms & 113.5 tok/s & 4.40$\times$ \\
\bottomrule
\end{tabular}}
\vspace{-2ex}
\label{tab:main_8b1b_latency}
\end{table}
% \vspace{-2ex}

\method reduces response time from 39.80 s to 9.13 s, a 4.36$\times$ speedup over \llamam-8B-only inference. 
Throughput increases from 25.81 tok/s to 113.5 tok/s, a 4.40$\times$ gain, because decode follows the \llamas-1B anchor. 
The response-time speedup and throughput gain are close because this workload is decode-dominated.

For energy calculation, we use the phase-specific power model from Splitwise~\citep{splitwise2024}. 
The prompt phase is modeled as compute-bound and uses 95\% of machine TDP, while decode is modeled as memory-bandwidth-bound and uses 45\% of aggregated GPU TDP:
% \vspace{-2ex}
\begin{equation}
\vspace{-1ex}
\resizebox{0.6\columnwidth}{!}{$
P_p = 2280\,\mathrm{W},  P_d = 1080\,\mathrm{W}.
\vspace{-2ex}
$}
\end{equation}

Energy per request drops from 11.98 Wh (i.e., watt-hour) for \llamam-8B-only inference to 2.78 Wh for \method, corresponding to a 4.31$\times$ reduction. 
 This shows \method preserves most of the energy efficiency of the small decode path while still using the large anchor for prompt processing.
\vspace{-1ex}
\begin{table}[!ht]
\centering\small
\caption{Energy results for \llamam-8B$\rightarrow$\llamas-1B, 1024/1024 tokens.}
\vspace{-2ex}
\resizebox{0.99\columnwidth}{!}{%
\begin{tabular}{lcccc}
\toprule
Model & Total energy & Energy/1K tok. & Tokens/Wh & Eng. Imp. \\
\midrule
Llama-1B         &  2.72 Wh &  2.66 Wh & 375.9 & 4.40$\times$ \\
Llama-8B         & 11.98 Wh & 11.70 Wh &  85.46 & baseline \\
8B$\rightarrow$1B  &  2.78 Wh &  2.71 Wh & 368.7 & 4.31$\times$ \\
\bottomrule
\end{tabular}}
\vspace{-2ex}
\label{tab:main_8b1b_energy}
\end{table}
\vspace{-2ex}
\subsection{Sensitivity to model-size gap}
\label{sec:model_gap}
\vspace{-1ex}
The benefit of \method is more pronounced when there is a large gap between the \prefillanchor and \decodeanchor, especially for output-heavy workloads. As it is shown in Table~\ref{tab:model_gap}, the benefit can go from 1.75$\times$ (1.76$\times$) to 33.86$\times$ (27.5$\times$) in terms of response time and energy savings.

\vspace{-2ex}
\begin{table}[!ht]
\centering
\small
\caption{Sensitivity to the model-size gap.}
\vspace{-2ex}
\resizebox{0.99\columnwidth}{!}{%
\begin{tabular}{lccc}
\toprule
Pair & Response speedup & Energy savings \\
\midrule
Llama-13B$\rightarrow$7B & 1.75$\times$ & 1.76$\times$ \\
Qwen-14B$\rightarrow$0.5B & 10.41$\times$ & 8.21$\times$ \\
Llama-70B$\rightarrow$1B & 33.86$\times$ & 27.5$\times$ \\
\bottomrule
\end{tabular}
}
\label{tab:model_gap}
\end{table}

\vspace{-2ex}
\section{Conclusion}
\label{sec:conclusion}
\vspace{-1ex}

We present \method, a multi-model LLM serving approach that uses a larger anchor for prefill and a smaller anchor for decode through a learned KV cache handoff.
This paper brings new insight to the efficient ML community.
Instead of forcing the entire request to run on one model, \method translates the prompt cache from the prefill-anchor space into the decode-anchor space, allowing the smaller model to handle the autoregressive steps with higher accuracy.
\method achieves the efficiency of the smaller model while exceeding its accuracy in every setting by up to 8.4\% relative, and closes up to 80\% of the gap to the larger model. On one A40 GPU, \llamam-8B$\rightarrow$\llamas-1B achieves 3.0$\times$ faster requests than \llamam-8B alone. \method improves the hardware and system solutions that separate
prefill and decode phases across GPUs. Our simulation results show that it reduces response time and reduces energy consumption compared to the larger model by 4.36$\times$ and 4.31$\times$ (\llamam-8B$\rightarrow$\llamas-1B), and up to 33.86$\times$ and 27.5$\times$ (\llamam-70B$\rightarrow$\llamas-1B).

\section*{Acknowledgment}
This material is based upon work supported in part by CoCoSys, one of seven centers in JUMP 2.0, a Semiconductor Research Corporation (SRC) program sponsored by DARPA, as well as sponsored research awards from Cisco Research and the National Science Foundation (NSF) under Grant Number CNS-2420977. 
Disclaimer: Any opinions, findings, conclusions, or recommendations expressed in this material are those of the authors and do not necessarily reflect the views of the National Science Foundation.

\bibliographystyle{latex/mlsys2025}
\bibliography{latex/custom}

\clearpage
\appendix
\section{Training Algorithms}
\label{sec:app_method_algorithms}

This section provides the algorithm-level training
procedures for \method. The main paper focuses on the mathematical
objective and the high-level serving recipe, while the algorithms
below spell out the execution order used in training and evaluation in
detail. Throughout the appendix, we follow the main-paper notation:
$M_p$ is the prefill anchor, $M_d$ is the decode anchor, $L_p$ and
$L_d$ are their layer counts, $\mathcal{C}^{p}$ is the source prompt
cache, $\mathcal{C}^{d}$ is the decode-anchor target cache, and
$\tilde{\mathcal{C}}^{d}$ is the stitched cache. We use $j$ for
decode-anchor layers and $i$ for prefill-anchor layers. When an
appendix equation discusses a generic key or value vector, we use $Z$
to avoid overloading $\mathcal{C}$, which denotes the full KV cache in
the main paper; $\pi(\ell)$ denotes the source layer in the
\prefillanchor used for target layer $\ell$ in the \decodeanchor.

\begin{algorithm}[!ht]
\small
\caption{Training \method}
\label{alg:training}
\begin{algorithmic}[1]
\REQUIRE Prefill anchor $M_p$, decode anchor $M_d$, dataset $\mathcal D$,
window size $W$, aggregation policy, KV-stitching module $G_\psi$,
loss weights $\lambda_{\mathrm{cos}},\lambda_{\mathrm{KL}},\lambda_{\mathrm{rec}}$
\STATE Tokenize $\mathcal D$ with the shared tokenizer and obtain train/validation/test splits.
\STATE Construct windows $\{\mathcal N_j\}_{j=1}^{L_d}$ using relative-depth centers $c_j$.
\STATE Initialize separate key/value stitching layers and, for weighted sums, mixing logits $\beta_j^K,\beta_j^V$.
\STATE Freeze the base parameters of $M_p$ and $M_d$.
\STATE Define $G_\psi$ to undo $M_p$'s RoPE on keys, aggregate source layers over $\mathcal N_j$, stitch keys and values, and apply $M_d$'s RoPE to output keys; values are not rotated.
\STATE Initialize the affine stitching projections by regularized least squares on paired source/target caches from training examples.
\IF{$\lambda_{\mathrm{rec}}>0$}
    \STATE Initialize an auxiliary linear decoder by least squares to reconstruct source caches from stitched caches.
\ENDIF
\STATE \textbf{Stage 1: Cache alignment and distillation}
\FOR{each training batch $(x,y)$}
    \STATE Run $M_p$ and $M_d$ on prompt $x$ to obtain source cache $\mathcal C^p$ and native target cache $\mathcal C^d$.
    \STATE $\widetilde{\mathcal C}^d\gets G_\psi(\mathcal C^p)$.
    \STATE Compute $\mathcal L_{\mathrm{align}}(\widetilde{\mathcal C}^d,\mathcal C^d)$ from MSE and $\lambda_{\mathrm{cos}}$-weighted cosine dissimilarity between corresponding key/value tensors.
    \IF{$\lambda_{\mathrm{rec}}>0$}
        \STATE Compute $\mathcal L_{\mathrm{rec}}$ by reconstructing source caches at window centers from $\widetilde{\mathcal C}^d$.
    \ENDIF
    \STATE Teacher-force reference answer $y$ through $M_p$ with $\mathcal C^p$ and $M_d$ with $\widetilde{\mathcal C}^d$.
    \STATE Compute $\mathcal L_{\mathrm{KL}}$ from $M_p$'s to stitched $M_d$'s distributions on valid answer tokens.
    \STATE Update $\psi$, and the auxiliary decoder if used, with $\mathcal L_{\mathrm{align}}+\lambda_{\mathrm{KL}}\mathcal L_{\mathrm{KL}}+\lambda_{\mathrm{rec}}\mathcal L_{\mathrm{rec}}$ (take $\mathcal L_{\mathrm{rec}}=0$ when disabled); keep both anchors frozen.
\ENDFOR
\STATE Save the Stage-1 stitch $G_\psi$, discard the auxiliary decoder if used, and record validation task score $s_1$.
\STATE \textbf{Stage 2: Task training}
\STATE Insert LoRA adapters $\eta$ into $M_d$; keep $M_p$ and both anchors' base parameters frozen, and keep $G_\psi$ trainable.
\STATE Choose the target $y$ for each $x$: the reference answer, or the answer that $M_p$ generates for $x$ (sequence-level distillation).
\FOR{each training batch $(x,y)$}
    \STATE Run $M_p$ on $x$ to obtain $\mathcal C^p$ and its first-answer-token logits.
    \STATE $\widetilde{\mathcal C}^d\gets G_\psi(\mathcal C^p)$.
    \STATE Teacher-force $y$ through $M_d$ with $\widetilde{\mathcal C}^d$ and adapters $\eta$ to predict the remaining answer tokens.
    \STATE Compute $\mathcal L_{\mathrm{task}}$ as cross-entropy against $y$, using $M_p$'s first-token logits and $M_d$'s remaining-token logits.
    \STATE Update $\psi$ and $\eta$ with $\mathcal L_{\mathrm{task}}$, optionally adding $\mathcal L_{\mathrm{KL}}$ with weight 0.5.
    \STATE Evaluate validation loss at the configured interval.
\ENDFOR
\STATE Restore the Stage-2 checkpoint with the lowest validation loss, including $G_\psi$ and $\eta$; record its validation task score $s_2$.
\STATE Report the Stage-1 checkpoint if $s_1\geq s_2$; otherwise report the Stage-2 checkpoint.
\end{algorithmic}
\end{algorithm}

\section{Dataset Details}
\label{sec:app_datasets}

Table~\ref{tab:app_datasets} lists, for each dataset, the prompt truncation length, the average prompt length, the average output length of the fine-tuned \llamas-1B, and the number of test examples.
The CoQA test set is hidden, so we use its official validation set for testing and hold out 200 training stories for validation.

\begin{table}[!ht]
\centering
\small
\caption{Dataset statistics. Max in: prompt truncation length; Avg.\ in: average prompt length; Avg.\ out: average output length of the fine-tuned \llamas-1B; Test: number of test examples. Lengths are in Llama-3 tokens.}
\vspace{-2ex}
\setlength{\tabcolsep}{4pt}
\begin{tabular}{lrrrr}
\toprule
Dataset & Max in & Avg.\ in & Avg.\ out & Test \\
\midrule
NarrativeQA & 2K & 821 & 6 & 10{,}557 \\
CoQA & 2K & 507 & 4 & 7{,}983 \\
SAMSum & 1K & 160 & 21 & 819 \\
Multi-News & 4K & 1{,}962 & 271 & 1{,}000 \\
GovReport & 8K & 6{,}734 & 499 & 973 \\
\bottomrule
\end{tabular}
\label{tab:app_datasets}
\end{table}

\section{Selected Training Configurations}
\label{sec:app_recipes}

Table~\ref{tab:recipes} lists the training stages selected on the
validation split for each model pair and dataset. ``S1'' means that
only the first stage is used, so both anchors are frozen and identical
to the fine-tuned baselines. ``S1+S2'' adds the second stage with
LoRA adapters on the \decodeanchor.
The ``Window'' column gives how the KV-stitching module combines the window of prefill layers for each decode layer (Section~\ref{sec:weighted_layer_mapping}): ``Sum'' is the weighted sum of Eq.~\eqref{eq:weighted_cache_mix}, and ``Concat'' is the concatenation variant.
All Llama settings use the weighted sum, and all Qwen settings use concatenation.

\begin{table}[!ht]
\centering
\small
\caption{Training stages and window combination selected on the validation split for each setting.}
\vspace{-2ex}
\begin{tabular}{llll}
\toprule
Dataset & Pair & Stages & Window \\
\midrule
NarrativeQA & Llama 3B$\rightarrow$1B & S1 & Sum \\
 & Llama 8B$\rightarrow$1B & S1 & Sum \\
 & Llama 8B$\rightarrow$3B & S1 & Sum \\
 & Qwen 3B$\rightarrow$0.5B & S1+S2 & Concat \\
 & Qwen 7B$\rightarrow$0.5B & S1+S2 & Concat \\
 & Qwen 7B$\rightarrow$3B & S1 & Concat \\
\midrule
CoQA & Llama 3B$\rightarrow$1B & S1+S2 & Sum \\
 & Llama 8B$\rightarrow$1B & S1 & Sum \\
 & Llama 8B$\rightarrow$3B & S1 & Sum \\
 & Qwen 3B$\rightarrow$0.5B & S1+S2 & Concat \\
 & Qwen 7B$\rightarrow$0.5B & S1+S2 & Concat \\
 & Qwen 7B$\rightarrow$3B & S1 & Concat \\
\midrule
SAMSum & Llama 3B$\rightarrow$1B & S1+S2 & Sum \\
 & Llama 8B$\rightarrow$1B & S1+S2 & Sum \\
 & Llama 8B$\rightarrow$3B & S1 & Sum \\
 & Qwen 3B$\rightarrow$0.5B & S1+S2 & Concat \\
 & Qwen 7B$\rightarrow$0.5B & S1+S2 & Concat \\
 & Qwen 7B$\rightarrow$3B & S1+S2 & Concat \\
\midrule
GovReport & Llama 3B$\rightarrow$1B & S1 & Sum \\
 & Llama 8B$\rightarrow$1B & S1 & Sum \\
 & Llama 8B$\rightarrow$3B & S1 & Sum \\
 & Qwen 3B$\rightarrow$0.5B & S1+S2 & Concat \\
 & Qwen 7B$\rightarrow$0.5B & S1+S2 & Concat \\
 & Qwen 7B$\rightarrow$3B & S1 & Concat \\
\midrule
Multi-News & Llama 3B$\rightarrow$1B & S1+S2 & Sum \\
 & Llama 8B$\rightarrow$1B & S1+S2 & Sum \\
 & Llama 8B$\rightarrow$3B & S1+S2 & Sum \\
 & Qwen 3B$\rightarrow$0.5B & S1+S2 & Concat \\
 & Qwen 7B$\rightarrow$0.5B & S1+S2 & Concat \\
 & Qwen 7B$\rightarrow$3B & S1+S2 & Concat \\
\bottomrule
\end{tabular}
\label{tab:recipes}
\end{table}
\vspace{-2ex}

\section{Additional ML Analysis and Ablation Details}
\label{sec:appendix_ablation}
This appendix provides additional details on the design choices
explored while developing \method. We focus on the components that
directly affect the learned KV handoff between $M_p$ and $M_d$: layer
mapping, stitching-layer design, alignment losses, optimization
schedules, gradient behavior, nonlinear transformations, residual
transformations, and teacher-guided warmup.

\subsection{Static layer mapping strategies}
\label{sec:app_static_mapping}

A KV handoff must decide which source layer from $M_p$ should provide
the cache used by each target layer in $M_d$. We first consider static
mapping policies. These policies are simple and low overhead, but they
may fail when the two anchors encode similar information at different
depths. 

We use $j \in \{1,\ldots,L_d\}$ for decode-anchor layers and $i \in
\{1,\ldots,L_p\}$ for prefill-anchor layers.

\subsubsection{Ordered mapping}
Ordered mapping maps decode layer $j$ in $M_d$ to the source layer
with the same relative position in $M_p$:
\begin{equation}
\pi_{\mathrm{ord}}(j) =
\operatorname{clip}(j,1,L_p).
\end{equation}
This policy is useful when the two anchors have similar depth and
similar layer-wise representation structure. However, it can lose
information when the two models have different numbers of layers or
when semantic information appears at different depths.

\subsubsection{Shifted mapping}
Shifted mapping maps decode layer $j$ in $M_d$ to a source layer with
a fixed offset:
\begin{equation}
\pi_{\mathrm{shift}}(j) =
\operatorname{clip}(j+\delta,1,L_p),
\end{equation}
where $\delta$ is a fixed integer offset. This strategy is useful when
representations in the larger prefill anchor appear later or earlier
than the corresponding representations in the smaller decode anchor.
The main limitation is that a single offset may not be optimal for all
layers or all tasks.

\subsubsection{Grouped mapping}
Grouped mapping divides the $L_p$ layers of $M_p$ into $L_d$
contiguous groups and maps the last layer of each group to one layer
of $M_d$. For decode layer $j$, the corresponding source group is
\begin{equation}
\mathcal{G}_{j}
=
\left\{
i:
\left\lfloor \frac{(j-1)L_p}{L_d} \right\rfloor + 1
\leq i \leq
\left\lfloor \frac{jL_p}{L_d} \right\rfloor
\right\}.
\end{equation}
The selected source layer is the last layer in this group:
\begin{equation}
\pi_{\mathrm{grp}}(j)
=
\max \mathcal{G}_{j}
=
\left\lfloor \frac{jL_p}{L_d} \right\rfloor .
\label{eq:app_grouped_mapping}
\end{equation}
In implementation, we clip $\pi_{\mathrm{grp}}(j)$ to the valid range
$[1,L_p]$. This policy is useful when deeper layers inside each group
contain aggregated information from the preceding layers. Its
limitation is that it compresses a group of source layers into one
selected layer, which may discard useful layer-specific information.

Table~\ref{tab:app_static_mapping} compares the accuracy of static
layer mapping methods against the proposed weighted-layer mapping. As
shown, weighted-layer mapping outperforms other approaches.

\begin{table}[!ht]
\centering
\small
\caption{Layer-mapping policies for the \llamas-3B$\rightarrow$\llamas-1B pair on NarrativeQA. Each policy trains only the first stage for one epoch without least-squares initialization, which supports only the weighted policy.}
\vspace{-2ex}
\setlength{\tabcolsep}{6pt}
\begin{tabular}{lccc}
\toprule
Layer mapping policy & F1 & ROUGE-L & BLEU \\
\midrule
Ordered mapping  & 53.8 & 55.7 & 4.4 \\
Shifted mapping  & 59.9 & 61.5 & 5.3 \\
Grouped mapping  & 60.4 & 62.0 & 5.7 \\
Weighted-layer mapping & \textbf{63.2} & \textbf{65.2} & \textbf{6.3} \\
\bottomrule
\end{tabular}
\label{tab:app_static_mapping}
\end{table}

\subsection{KV cache stitching strategies}
\label{sec:app_stitching_family}

After selecting the source cache for each target layer, \method must
transform it into the representation space expected by $M_d$. We
compare five broad classes of KV cache stitching strategies: linear,
MLP, convolutional transformations, bottleneck stitching, and residual
stitching. Each choice offers a different balance between simplicity,
expressiveness, fitting capacity, and optimization stability.
Table~\ref{tab:app_stitching_family} compares these approaches.

\subsubsection{Linear stitching}
A linear stitch layer treats each token position independently and
only changes the feature dimension. Let $Z$ denote either a key vector
or a value vector from the selected source layer. The linear stitch is
\begin{equation}
\tilde{Z}^{d}_{j,t}
=
P_{j}Z^{p}_{\pi(j),t}
+
b_{j},
\label{eq:app_linear_stitch}
\end{equation}
where $j$ is the decode-anchor layer and $t$ is the token position.
This design is fast, simple, and low overhead. In our early
experiments, simple linear layers were reasonably effective to capture
useful semantic alignment. However, a single linear projection cannot
correct nonlinear mismatches between the source and target
representations, which can limit its effectiveness when the two
anchors differ substantially in scale or internal structure.

\subsubsection{MLP stitching}
An MLP stitch layer adds a hidden projection and a nonlinearity:
\begin{equation}
\tilde{Z}^{d}_{j,t}
=
D_{j}
\sigma
\left(
U_{j}Z^{p}_{\pi(j),t}
+
b^{(1)}_{j}
\right)
+
b^{(2)}_{j}.
\label{eq:app_mlp_stitch}
\end{equation}
In the default setup, the hidden dimension is
\begin{equation}
d_{\mathrm{hidden}}=\frac{d_{\mathrm{in}} + d_{\mathrm{out}}}{2}.
\label{eq:app_mlp_hidden}
\end{equation}
The benefit of this design is that it can model nonlinear
relationships between the two KV spaces. However, the nonlinearity can
hurt transformation learning at the beginning of training because the
model must learn both alignment and nonlinear correction at the same
time. It also makes the optimization landscape harder, which can make
it more difficult to find a good minimum. Our results suggest that MLP
stitching should be used carefully and may require warmup, residual
scaling, or stronger regularization. These limitations motivated the
residual nonlinear form in Eq.~\ref{eq:nonlinear_stitch}, which
preserves the simplicity and stability of the linear stitch while
introducing the nonlinear branch gradually through the learned
residual scale $\alpha_j$.

\subsubsection{Convolutional stitching}
A convolutional stitch layer applies local transformations across
nearby token positions:
\begin{equation}
\tilde{Z}^{d}_{j,t}
=
\sum_{\Delta=-q}^{q}
W_{j,\Delta}
Z^{p}_{\pi(j),t+\Delta}.
\label{eq:app_conv_stitch}
\end{equation}
This can capture local dependencies across adjacent tokens in the KV
 cache. $\Delta$ is the relative token-position offset around position
 $t$, and $q$ is the convolution radius. 
Such locality may be useful because attention patterns often have
strong local structure. However, convolutional stitching adds
sequence-level computation and may increase inference overhead.

\subsubsection{Bottleneck stitching}
\label{sec:app_bottleneck}

A simple linear stitch layer may be too weak when the source and
target KV spaces have a nonlinear relationship. To address this, we
explored bottleneck stitching. The transformation first projects the
input to a smaller hidden dimension, applies normalization and
nonlinearity, and then projects back to the target dimension:
\begin{equation}
\tilde{Z}^{d}_{j,t}
=
W_{\mathrm{up},j}
\sigma
\left(
\mathrm{LN}
\left(
W_{\mathrm{down},j}
Z^{p}_{\pi(j),t}
\right)
\right).
\label{eq:app_bottleneck}
\end{equation}
The nonlinearity makes this module more expressive than a pure linear
map, and the intermediate LayerNorm changes the geometry of the
representation before the up-projection. Although bottleneck stitching
increases capacity, it did not consistently improve downstream
performance. In our experiments, it started to overfit early, even
when the dataset size was increased. This indicates that adding
nonlinear capacity alone is not enough. The stitch layer must improve
generation behavior, not only reduce cache-alignment loss. Bottleneck
stitching therefore provides a useful stress test: it is more
expressive than a linear map, but that extra expressiveness used
naively can overfit early and may not translate into better generation
quality.

\subsubsection{Residual stitching}
\label{sec:app_residual}

Residual stitching keeps a stable linear path and adds a nonlinear
correction branch:
\begin{equation}
\tilde{Z}^{d}_{j,t}
=
P_{j}Z^{p}_{\pi(j),t}
+
\alpha_{j}
g_{j}
\left(
Z^{p}_{\pi(j),t}
\right),
\label{eq:app_residual}
\end{equation}
where $P_{j}$ is the linear path, $g_{j}(\cdot)$ is the residual
branch, and $\alpha_{j}$ controls the residual strength. This design
is motivated by the observation that a stable linear map is useful for
early training, while a small nonlinear branch may correct difficult
layers.

We considered three residual designs. First, a nonlinear layer with a
linear residual path. This was unstable because the randomly
initialized nonlinear branch became responsible for a large part of
the mapping early in training. Second, a linear layer with a linear
residual path. This adds parameters but no real expressiveness,
because the sum of two linear maps is still a linear map. Third, a
linear path with a nonlinear residual branch. This is the most
promising design because the linear path provides stable alignment
while the nonlinear branch learns corrections.

\begin{table}[!ht]
\centering
\small
\caption{Test ROUGE-L of KV cache stitching designs for \llamam-8B$\rightarrow$\llamas-1B on NarrativeQA. Residual stitching is the design of \method (Section~\ref{sec:kv_layer}).}
\setlength{\tabcolsep}{6pt}
\begin{tabular}{lc}
\toprule
Stitching-layer design  & ROUGE-L \\
\midrule
Linear stitching               & 64.38 \\
MLP stitching                   & 64.83 \\
Convolutional stitching         & 63.86 \\
Bottleneck stitching      & 63.37 \\
Residual stitching (\method)       & \textbf{66.99} \\
\bottomrule
\end{tabular}
\label{tab:app_stitching_family}
\end{table}

\subsection{KV alignment loss}
\label{sec:app_alignment_loss}

\paragraph{Cosine similarity.}The main paper defines the alignment
objective between the stitched cache $\tilde{\mathcal{C}}^{d}$ and
the decode-anchor target cache $\mathcal{C}^{d}$. Here, we expand the
element-wise form. Let $Q \in \{K,V\}$ denote either key or value
states. Directly minimizing mean squared error encourages numerical
closeness:
\begin{equation}
\resizebox{0.86\columnwidth}{!}{$
\mathcal{L}_{\mathrm{MSE}}
=
\frac{1}{|\mathcal{B}|}
\sum_{\mathbf{x}\in \mathcal{B}}
\sum_{j,t}
\sum_{Q\in\{K,V\}}
\left\|
\tilde{Q}^{d}_{j,t}(\mathbf{x})
-
Q^{d}_{j,t}(\mathbf{x})
\right\|_2^2 .
\label{eq:app_mse_loss}
$}
\end{equation}
However, MSE alone may not fully capture semantic alignment. Two
vectors can be close in direction but different in scale, or close in
value but less aligned in the direction that matters for attention.

To address this, we add a cosine-similarity term:
\begin{equation}
\resizebox{0.85\columnwidth}{!}{$
\resizebox{0.90\columnwidth}{!}{$
\displaystyle
\mathcal{L}_{\mathrm{cos}}
=
\frac{1}{|\mathcal{B}|}
\sum_{\mathbf{x}\in \mathcal{B}}
\sum_{j,t}
\sum_{Q\in\{K,V\}}
\left(
1 -
\frac{
\left\langle
\tilde{Q}^{d}_{j,t}(\mathbf{x}),
Q^{d}_{j,t}(\mathbf{x})
\right\rangle
}{
\left\|
\tilde{Q}^{d}_{j,t}(\mathbf{x})
\right\|_2
\left\|
Q^{d}_{j,t}(\mathbf{x})
\right\|_2
}
\right).
$}
\label{eq:app_cos_loss}
$}
\end{equation}
The total alignment loss is
\begin{equation}
\mathcal{L}_{\mathrm{align}}
=
\mathcal{L}_{\mathrm{MSE}}
+
\lambda_{\mathrm{cos}}
\mathcal{L}_{\mathrm{cos}},
\label{eq:app_total_align_loss}
\end{equation}
where $\lambda_{\mathrm{cos}}$ is the same cosine-loss weight used in
Eq.~\ref{eq:align_loss}.

Empirically, adding the cosine term improved text-quality metrics such
as ROUGE and BLEU compared with an MSE-only baseline. This suggests
that the stitch layer should preserve not only the raw cache values,
but also the direction of the cache vectors. In other words, the
cosine term encourages semantic-direction alignment, which can improve
downstream generation metrics even when MSE alone appears to fit the
cache values.

\paragraph{Cosine learning rate decay.} We also explored cosine
learning-rate decay for training the stitch module. To avoid
overloading the adapter parameter notation $\eta$ from the main paper,
we denote the learning rate at step $t$ by $\xi_t$. Unlike linear
decay, cosine decay decreases the learning rate smoothly from a
maximum value toward a small value:
\begin{equation}
\resizebox{0.85\columnwidth}{!}{$
\displaystyle
\xi_t
=
\xi_{\min}
+
\frac{1}{2}
(\xi_{\max} - \xi_{\min})
\left(
1 + \cos\left(\frac{\pi t}{T}\right)
\right).
$}
\label{eq:app_cos_lr}
\end{equation}
Here, $t$ is the current step and $T$ is the total number of training
steps. Cosine decay can also be used with restarts, which periodically
reset the learning rate and may help escape poor local minima.

This schedule is useful for small trainable modules such as LoRA
adapters and KV cache stitching layers. It avoids dropping the
learning rate too quickly, which can stall learning, while still
reducing the update size near the end of training. In our experiments,
the cosine-LR setup matched or exceeded the linear-decay baseline
across ROUGE and BLEU metrics, suggesting more stable text quality.
Smooth decay can therefore stabilize updates for small trainable
modules and improve generation quality, as summarized in
Table~\ref{tab:app_summary}.

\begin{table}[!ht]
\centering
\small
\caption{Effect of adding a cosine-similarity term to the MSE alignment loss (\llama-3B$\rightarrow$\llama-1B, NarrativeQA), in preliminary experiments.}
\setlength{\tabcolsep}{6pt}
\begin{tabular}{lcc}
\toprule
Alignment loss & ROUGE-1 & ROUGE-2 \\
\midrule
MSE only       & 0.130 & 0.085 \\
MSE + cosine   & 0.170 & 0.105 \\
\bottomrule
\end{tabular}
\label{tab:app_alignment_loss}
\end{table}

\subsection{Gradient analysis and layer compatibility}
\label{sec:app_gradient_analysis}

Our gradient and similarity analysis reveals three important issues.

First, some decode-anchor layers have weak linear correspondence with
all candidate prefill-anchor layers. In our analysis,
 several layers appear to have KV representations that are not well
explained by a simple linear relationship to any source layer. For
such layers, changing the mapping policy alone may not be enough,
because the mismatch is likely nonlinear.

Second, we observe gradient imbalance across layers. Some layers
receive much smaller gradients than others.
This can cause parts of the stitch module to under-train while other
parts dominate optimization. This motivates layer-wise learning-rate
schedules or residual branches that give difficult layers more useful
capacity.

Third, increasing transformation capacity does not always improve
generation quality. In some cases, the stitch layer can fit the
training objective more aggressively, but downstream performance does
not improve. This suggests that stronger reconstruction or more
parameters are not always equivalent to better task behavior. The
residual nonlinear form in Eq.~\ref{eq:nonlinear_stitch} is motivated
by this observation: it adds controlled nonlinear capacity through a
scaled residual branch while retaining a stable linear path. 

Overall, this analysis shows that some layers have weak linear
correspondence, some layers receive much smaller gradients, and
higher-capacity transformations do not always improve task quality.
These observations motivate the later exploration of layer-wise
learning rates, bottleneck transformations, and residual
transformations.

\subsection{Layer-wise learning rate and LoRA rank}
\label{sec:app_layerwise_lr_lora}

Because different stitch layers learn at different speeds, we
considered layer-wise learning-rate schedules. The main idea is to
assign higher learning rates to layers with weaker gradients and lower
learning rates to layers that already learn quickly. To avoid conflict
with the adapter-parameter notation $\eta$ in the main paper, we
denote learning rates by $\xi$:
\begin{equation}
\xi_{j}
=
\xi_0 \cdot s_{j},
\label{eq:app_layerwise_lr}
\end{equation}
where $s_{j}$ is a layer-specific scaling factor.

We also explored increasing the LoRA rank and scaling factor of the second stage.
For \llamas-3B$\rightarrow$\llamas-1B on NarrativeQA, rank $16$ with $\alpha=64$ reaches the same ROUGE-L as rank $8$ with $\alpha=16$ (65.66; Table~\ref{tab:app_layerwise_lr_lora}), so the additional adapter capacity does not improve task quality.
In some cases, the smaller anchor's standalone behavior degraded, suggesting that additional adapter capacity must be used carefully.
Table~\ref{tab:app_lr_lora} summarizes our observations from the layer-wise learning-rate and LoRA-rank ablations.

\begin{table}[!ht]
\centering
\small
\caption{Summary of observations from layer-wise learning-rate and LoRA-rank ablations.}
\setlength{\tabcolsep}{5pt}
\begin{tabular}{p{0.35\linewidth}p{0.55\linewidth}}
\toprule
Design choice & Observation \\
\midrule
Layer-wise LR & Helps compensate for gradient imbalance across stitch
layers. \\
Higher LoRA rank & Does not improve ROUGE-L over rank 8
(Table~\ref{tab:app_layerwise_lr_lora}). \\
Higher LoRA scaling & Adds adaptation strength, but may destabilize
the smaller anchor. \\
\bottomrule
\end{tabular}
\label{tab:app_lr_lora}
\end{table}

\begin{table}[!ht]
\centering
\small
\caption{LoRA-rank ablation of the second stage (\llamas-3B$\rightarrow$\llamas-1B, NarrativeQA, ROUGE-L).}
\setlength{\tabcolsep}{6pt}
\begin{tabular}{lc}
\toprule
Training configuration & ROUGE-L \\
\midrule
Uniform LR $5\times10^{-5}$, LoRA $r=8$, $\alpha=16$  & 65.66 \\
Uniform LR $5\times10^{-5}$, LoRA $r=16$, $\alpha=64$ & 65.66 \\
\bottomrule
\end{tabular}
\label{tab:app_layerwise_lr_lora}
\end{table}

\subsection{Teacher-Guided Warmup and Knowledge Distillation}
\label{sec:app_kd}

A key issue in training the stitch layer is the mismatch between the
warmup objective and the final task objective. During warmup, the
model is optimized to align transformed KV caches with target KV
caches. During actual training and evaluation, however, the model is
judged by generation quality. A lower KV-alignment loss can provide a
better starting point for task training, but it does not always imply
better ROUGE, BLEU, or task loss. 

To reduce this mismatch, we explored teacher-guided warmup. The
prefill anchor acts as a teacher, and the warmup loss combines KV
transformation loss with a teacher-forcing task loss:
\begin{equation}
\mathcal{L}_{\mathrm{warmup}}
=
\mu_{\mathrm{KV}}
\mathcal{L}_{\mathrm{KV}}
+
\mu_{\mathrm{teacher}}
\mathcal{L}_{\mathrm{teacher}},
\label{eq:app_kd_loss}
\end{equation}
where $\mathcal{L}_{\mathrm{KV}}$ is the KV transformation loss,
$\mathcal{L}_{\mathrm{teacher}}$ encourages the stitched model to
follow the teacher model's behavior on the downstream task, and
$\mu_{\mathrm{KV}}$ and $\mu_{\mathrm{teacher}}$ control the two
terms. We use $\mu$ for these warmup weights to avoid conflict with
the residual scale $\alpha_j$ and the weighted-layer mapping logits
$\beta^K_j,\beta^V_j$ used in the main paper.

Table~\ref{tab:app_kd} compares warmup strategies before the second stage for \llamas-3B$\rightarrow$\llamas-1B on NarrativeQA.
Without KV warmup, the second stage reaches only 23.90 and 27.78 ROUGE-L at learning rates of $5\times10^{-5}$ and $2\times10^{-4}$.
After KV-alignment warmup, the same second stage reaches 65.36, so aligning the cache before task training is essential.
Adding the teacher term with a KL weight of 0.1 to the warmup raises the score further, to 65.66.
We also observed that value-cache gradients could diminish much faster than key-cache gradients.

\begin{table}[!ht]
\centering
\small
\caption{Warmup ablation (\llamas-3B$\rightarrow$\llamas-1B, NarrativeQA, ROUGE-L after the second stage). LR is the learning rate of the second stage. KV-alignment warmup is essential, and adding the teacher term to it gives the highest score.}
\setlength{\tabcolsep}{6pt}
\begin{tabular}{lcc}
\toprule
Warmup & LR & ROUGE-L \\
\midrule
None & $5\times10^{-5}$ & 23.90 \\
None & $2\times10^{-4}$ & 27.78 \\
KV alignment (KL weight 0) & $5\times10^{-5}$ & 65.36 \\
Teacher-guided (KL weight 0.1) & $5\times10^{-5}$ & \textbf{65.66} \\
\bottomrule
\end{tabular}
\label{tab:app_kd}
\end{table}

\subsection{Summary of layer mapping and KV cache stitching findings}
\label{sec:app_summary}

\begin{table*}[!h]
\centering
\small
\caption{Summary of ML analysis findings. The main lesson is that better KV cache reconstruction alone is not sufficient; the stitch layer must preserve downstream generation behavior while remaining lightweight and stable.}
\vspace{-1ex}
\setlength{\tabcolsep}{5pt}
\begin{tabular}{p{0.22\linewidth}p{0.34\linewidth}p{0.34\linewidth}}
\toprule
Design & Main benefit & Main limitation \\
\midrule
\multicolumn{3}{l}{\textit{Mapping Policies}} \\
\midrule
Ordered mapping & Simple and low overhead & Fails when model depths or
layer semantics differ \\
Shifted mapping & Accounts for depth offset between anchors & Requires
manually choosing a fixed offset per model pair \\
Grouped mapping & Compresses larger-anchor depth into smaller-anchor
depth & Can lose layer-specific information \\
Weighted-layer mapping & Learns a local mixture of source layers &
Adds parameters and needs regularization \\
\midrule
\multicolumn{3}{l}{\textit{Stitching Layer Designs}} \\
\midrule
Linear stitching & Simple, stable, and efficient & Weak for nonlinear
layer mismatch \\
MLP stitching & Adds nonlinear capacity & Harder early optimization \\
Convolutional stitching & Captures local token dependencies & Adds
sequence-level overhead \\
Bottleneck stitching & Adds nonlinear bottleneck capacity & Discards
geometric structure before nonlinear branch acts; no stable linear
path \\
Residual stitching & Keeps a stable linear path with nonlinear
correction & Requires careful residual scaling \\
\midrule
\multicolumn{3}{l}{\textit{Alignment Objectives}} \\
\midrule
MSE + cosine loss & Aligns both value and semantic direction &
Requires tuning the cosine weight \\
Teacher-guided warmup & Highest score after the second stage
(Table~\ref{tab:app_kd}) & Needs a KL weight on top of the alignment loss \\
\midrule
\multicolumn{3}{l}{\textit{Optimization}} \\
\midrule
Cosine LR decay & Stabilizes updates for small trainable modules &
Needs schedule tuning \\
Layer-wise LR & Addresses gradient imbalance & Tuning cost scales
linearly with $L_d$ \\
Higher LoRA rank & Improves fitting capacity & Does not always improve
task metrics \\
\bottomrule
\end{tabular}
\label{tab:app_summary}
\end{table*}

Overall, three key findings emerge from the ablation studies. First,
static layer mapping fails when anchor depths or layer semantics
differ, motivating the weighted-layer approach. Second, alignment loss
and generation quality can decouple: more expressive transformations
or lower KV cache MSE do not reliably improve downstream metrics,
because fitting cache values is not the same as preserving
attention-compatible geometry. Third, the residual nonlinear stitch
resolves the stability-expressiveness tradeoff by keeping a linear
path intact while introducing nonlinear correction gradually through
the scaled residual branch $\alpha_j$, a design that no other ablated
configuration achieves. Table~\ref{tab:app_summary} summarizes the
main lessons from ML analysis and ablation studies.

\section{Additional System Results and Sensitivity Studies}
\label{sec:appendix_sys_results-overview}

This section reports the additional system-side results that are not
included in the main text. We include the
\llamal-13B$\rightarrow$\llamal-7B results, the
\qwen-14B$\rightarrow$\qwen-0.5B results, and sensitivity studies for
scaling the number of requests, input length, and output length. The
details of the simulator system, which is built on top of the
Splitwise~\citep{splitwise2024} and workload configuration, are shown
in Table~\ref{tab:hw-setup}.
\begin{table}[!ht]
\centering
\small
\caption{Simulator system and workload configuration}
\vspace{-2ex}
\begin{tabular}{ll}
\toprule
Parameter & Value \\
\midrule
Hardware & 8 NVIDIA A40 GPUs \\
GPU TDP & 300 W per GPU \\
Aggregate GPU TDP & 2400 W \\
Precision & FP16 \\
Bytes per element & 2 bytes \\
Tensor parallelism & 1 \\
Arrival process & Exponential inter-arrival times \\
Arrival rate & 2.0 requests/s \\
Intra-machine (NVLink) & 100~GB/s \\
Inter-machine InfiniBand & 50~GB/s\\ 
\bottomrule
\end{tabular}
\vspace{-2ex}
\label{tab:hw-setup}
\end{table}

\subsection{\llamal-13B$\rightarrow$\llamal-7B results}
\label{sec:appendix_13b7b}

Table~\ref{tab:appendix_13b7b_latency} reports the
\llamal-13B$\rightarrow$\llamal-7B results for 1024 input tokens and
1024 output tokens, measured with intra-node NVLink for KV-cache
handoff. The larger 13B model performs prefill, while the 7B model
performs decode. Because decode dominates this workload, the response
time of \method is close to the 7B anchor and much faster than
\llamal-13B-only inference.

\vspace{1ex}
\begin{table}[!ht]
\centering
\small
\caption{\llamal-13B$\rightarrow$\llamal-7B latency and throughput
results for 200 requests with 1024 input and 1024 output tokens.}
\vspace{-2ex}
\resizebox{0.99\columnwidth}{!}{%
\begin{tabular}{lccccc}
\toprule
Model & Response & TTFT & TBT & Throughput & Improvement \\
\midrule
 Llama-7B   & 35.99 s & 112.35 ms & 35.04 ms & 28.45 tok/s &
 1.76$\times$ \\
 Llama-13B & 63.38 s & 202.89 ms & 61.69 ms & 16.16 tok/s & baseline
 \\
 13B$\rightarrow$7B & 36.14 s & 192.75 ms & 35.11 ms & 28.33 tok/s &
 1.75$\times$  \\
\bottomrule
\end{tabular}
} 
\label{tab:appendix_13b7b_latency}
\end{table}

Table~\ref{tab:appendix_13b7b_energy} reports the corresponding energy
results. The energy reduction mainly comes from replacing 13B decoding
with 7B decoding.

\begin{table}[!ht]
\centering
\small
\caption{\llamal-13B$\rightarrow$\llamal-7B energy results for 200
requests with 1024 input and 1024 output tokens.}
\vspace{-2ex}
\resizebox{0.99\columnwidth}{!}{%
\begin{tabular}{lcccc}
\toprule
Model & Total energy & Energy/1K tok. & Tokens/Wh & Eng. Imp. \\
\midrule
Llama-7B   & 10.83 Wh & 10.58 Wh & 94.52 & 1.76$\times$  \\
Llama-13B & 19.08 Wh & 18.63 Wh & 53.67 & baseline \\
13B$\rightarrow$7B & 10.91 Wh & 10.65 Wh & 93.88 & 1.75$\times$  \\
\bottomrule
\end{tabular}
} 
\label{tab:appendix_13b7b_energy}
\end{table}

\subsection{Scaling with request count for \llamal-13B$\rightarrow$\llamal-7B}
\label{sec:appendix_request_scaling_13b7b}

Our results show that \method is almost stable as the number of
requests increases. Table~\ref{tab:appendix_request_latency_13b7b}
shows response-time scaling from 200 to 1000 requests (queries). All
response times change by less than 0.10 s, which means the system
behavior is predictable as the number of requests increases.

\begin{table}[!ht]
\centering
\small
\caption{Response-time scaling for the
\llamal-13B$\rightarrow$\llamal-7B simulator result.}
\vspace{-2ex}
\resizebox{0.99\columnwidth}{!}{%
\begin{tabular}{lccc}
\toprule
Model & 200 requests & 500 requests & 1000 requests \\
\midrule
Llama-7B   & 35.99 s & 36.00 s & 36.02 s \\
Llamal-13B & 63.38 s & 63.45 s & 63.48 s \\
13B$\rightarrow$7B & 36.14 s & 36.17 s & 36.20 s \\
\bottomrule
\end{tabular}
} 
\label{tab:appendix_request_latency_13b7b}
\end{table}

Table~\ref{tab:appendix_request_ttft_13b7b} shows that the TTFT
advantage improves under higher load. The reason is that faster decode
reduces queue buildup, which helps first-token latency as the request
count increases. The TTFT advantage is reported as an absolute
reduction in milliseconds.

\begin{table}[!ht]
\centering
\small
\caption{TTFT scaling for the \llamal-13B$\rightarrow$\llamal-7B
simulator result (intra-node NVLink). The TTFT advantage grows with
request count because faster decode reduces queueing.}
\vspace{-2ex}
\resizebox{0.99\columnwidth}{!}{%
\begin{tabular}{lccc}
\toprule
Model & 200 requests & 500 requests & 1000 requests \\
\midrule
Llama-7B   & 112.35 ms & 115.97 ms & 117.11 ms \\
Llama-13B & 202.89 ms & 209.24 ms & 211.73 ms \\
13B$\rightarrow$7B & 192.75 ms & 197.20 ms & 198.43 ms \\
\midrule
\method vs.\ 13B & $-10.14$ ms & $-12.04$ ms & $-13.30$ ms \\
\bottomrule
\end{tabular}
} 
\label{tab:appendix_request_ttft_13b7b}
\end{table}

Energy is also stable across request counts, as shown in
Table~\ref{tab:appendix_request_energy_13b7b}.

\begin{table}[!ht]
\centering
\small
\caption{Energy stability for the \llamal-13B$\rightarrow$\llamal-7B
simulator result.}
\vspace{-2ex}
\resizebox{0.99\columnwidth}{!}{%
\begin{tabular}{lccc}
\toprule
Metric & 200 requests & 500 requests & 1000 requests \\
\midrule
\method energy/request & 10.91 Wh & 10.92 Wh & 10.93 Wh \\
13B energy/request     & 19.08 Wh & 19.11 Wh & 19.11 Wh \\
Energy improvement     & 1.75$\times$ & 1.75$\times$ & 1.75$\times$ \\
\method tokens/Wh      & 93.88 & 93.80 & 93.71 \\
Tokens/Wh gain vs.\ 13B & 1.75$\times$ & 1.75$\times$ & 1.75$\times$
\\
\bottomrule
\end{tabular}
} 
\label{tab:appendix_request_energy_13b7b}
\end{table}

\subsection{\qwen-14B$\rightarrow$\qwen-0.5B}
\label{sec:appendix_qwen14b05b}

Table~\ref{tab:appendix_qwen_latency} reports several quantitative
results in the setting \qwen-14B $\rightarrow$ \qwen-0.5B. These
results show how \method behaves when the decode model is much smaller
than the prefill model. The large model performs prefill once, while
the very small model handles the repeated decode phase. Because the
0.5B decode anchor has only 24 layers, its per-token decode time is
bounded below by per-layer kernel overhead, which limits how small TBT
can become.

\begin{table}[!ht]
\centering
\small
\caption{\qwen-14B$\rightarrow$\qwen-0.5B latency and throughput for
1024 input and 1024 output tokens.}
\vspace{-2ex}
\resizebox{0.99\columnwidth}{!}{%
\begin{tabular}{lccccc}
\toprule
Model & Response & TTFT & TBT & Throughput & Improvement \\
\midrule
Qwen-0.5B    &  8.20 s &  19.39 ms &  7.99 ms & 125.21 tok/s &
8.58$\times$  \\
Qwen-14B    & 70.31 s & 233.37 ms & 68.43 ms &  14.61 tok/s & baseline
\\
14B$\rightarrow$0.5B &  8.38 s & 197.14 ms &  7.99 ms & 125.21 tok/s &
8.39$\times$ \\
\bottomrule
\end{tabular}
} 
\label{tab:appendix_qwen_latency}
\end{table}

The response-time speedup is 8.39$\times$ over \qwen-14B-only
inference. This large gain comes from the large decode-model gap: the
0.5B decode anchor has much lower TBT than the 14B anchor.

\begin{table}[!ht]
\centering
\small
\caption{\qwen-14B$\rightarrow$\qwen-0.5B energy result for 1024
input and 1024 output tokens.}
\vspace{-1ex}
\resizebox{0.99\columnwidth}{!}{%
\begin{tabular}{lcccc}
\toprule
Model & Total energy & Energy/1K tok. & Tokens/Wh & Eng. Imp. \\
\midrule
Qwen-0.5B     &  2.47 Wh &  2.41 Wh & 415.28 & 8.59$\times$  \\
Qwen-14B     & 21.17 Wh & 20.67 Wh &  48.37 & baseline \\
14B$\rightarrow$0.5B &  2.58 Wh &  2.52 Wh & 397.15 & 8.21$\times$ \\
\bottomrule
\end{tabular}
} 
\label{tab:appendix_qwen_energy}
\end{table}

Energy per request decreases from 21.17 Wh to 2.58 Wh, an 8.21$\times$
reduction. The method remains slightly more expensive than 0.5B-only
inference because the prompt is still processed by the larger 14B
anchor.

\subsection{Sensitivity to input length}
\label{sec:appendix_input_length}

Table~\ref{tab:appendix_input_length} varies the input length while
keeping the output length fixed at 1024 tokens for the main
\llamam-8B$\rightarrow$\llamas-1B setting. As input length grows,
prefill becomes more important, so the speedup decreases. 
 Even with 16K input tokens, \method still provides a 3.74$\times$
response-time speedup because decode remains a major part of the 1024
output token workload.

\begin{table}[!ht]
\centering
\small
\caption{Input-length sensitivity for
\llamam-8B$\rightarrow$\llamas-1B with output length fixed at 1024
tokens. Longer inputs increase the prefill fraction, reducing speedup,
but \method remains substantially faster.}
\vspace{-2ex}
\resizebox{0.99\columnwidth}{!}{%
\begin{tabular}{lcccc}
\toprule
Input length & 8B res. & \method res. & Speedup & Energy imp. \\
\midrule
512   & 39.65 s &  9.06 s & 4.38$\times$ & 4.35$\times$ \\
1024  & 39.83 s &  9.14 s & 4.36$\times$ & 4.32$\times$ \\
2048  & 40.19 s &  9.30 s & 4.32$\times$ & 4.24$\times$ \\
4096  & 40.93 s &  9.65 s & 4.24$\times$ & 4.09$\times$ \\
16384 & 45.89 s & 12.28 s & 3.74$\times$ & 3.27$\times$ \\
\bottomrule
\end{tabular}
} 
\label{tab:appendix_input_length}
\end{table}

\subsection{Sensitivity to output length}
\label{sec:appendix_output_length}

Table~\ref{tab:appendix_output_length} varies the output length while
keeping the input length fixed at 1024 tokens for the main
\llamam-8B$\rightarrow$\llamas-1B setting. The relative speedup
saturates because both \llamam-8B-only inference and \method scale
almost linearly with output length once decode dominates, and the
ratio approaches the decode-speed ratio
TBT$_{\mathrm{8B}}$/TBT$_{\mathrm{1B}} = 4.40\times$. The absolute
latency and energy savings, however, continue to grow with output length.

\begin{table}[!ht]
\centering
\small
\caption{Output-length sensitivity for
\llamam-8B$\rightarrow$\llamas-1B with input length fixed at 1024
tokens. Improvements are measured against \llamam-8B-only inference.
The speedup ratio saturates because both systems scale linearly with
output length, but absolute latency and energy savings grow with the
number of generated tokens.}
\vspace{-1ex}
\resizebox{\columnwidth}{!}{%
\begin{tabular}{lccccccc}
\toprule
Output & 8B resp. & \method resp. &  Speedup & Tput gain &  Eng Imp.
\\
\midrule
512    &   19.98 s &    4.62 s &   4.32$\times$ & 4.40$\times$ &
4.24$\times$ \\
1024   &   39.83 s &    9.14 s &   4.36$\times$ & 4.40$\times$ &
4.32$\times$ \\
8192   &  317.74 s &   72.31 s &   4.39$\times$ & 4.40$\times$ &
4.39$\times$ \\
16384  &  635.35 s &  144.52 s & 4.40$\times$ & 4.40$\times$ &
4.39$\times$ \\
\bottomrule
\end{tabular}
} 
\label{tab:appendix_output_length}
\end{table}

\subsection{Summary of system-side findings}
\label{sec:appendix_system_summary}

The additional system results show that \method is most beneficial
when decode dominates the request and when the decode anchor is much
smaller than the prefill anchor. For the measured
\llamal-13B$\rightarrow$\llamal-7B setting, \method achieves
1.75$\times$ faster response time and 1.75$\times$ lower energy than
\llamal-13B-only inference. For the main
\llamam-8B$\rightarrow$\llamas-1B setting, \method achieves
4.36$\times$ faster response time and 4.32$\times$ lower energy at
1024 output tokens. For longer outputs, the relative speedup saturates
near the decode-speed ratio of 4.40$\times$, but the absolute savings
continue to grow. For the \qwen-14B$\rightarrow$\qwen-0.5B setting,
\method achieves a large gain in our study, with 8.39$\times$ faster
response time and 8.21$\times$ lower energy than \qwen-14B-only
inference. As we showed in the main paper, the advantage can go up to
33.86$\times$ and 27.5$\times$ in terms of response time and energy
savings, when using \llamal-70B for prefill and \llamas-1B for decode.

\providecommand{\scalesize}{0.95}
\section{Comparison with related work}
\label{app:related_comparison}

Table~\ref{tab:related_comparison} compares serving features,
cache-layer mapping, connector supervision, and evaluation scope
across related methods.
Here, the source is the model that produces the prompt cache, and the
receiver is the model that uses the cache for generation.
For \method{}, these are the larger prefill anchor and the smaller
decode anchor, respectively.

\begin{table*}[tbp]
\centering
\caption{Comparison of related approaches. \(\checkmark\): supported;
\(\times\): not supported in the cited recipe; \(\triangle\):
restricted or variant-dependent; ---: not applicable. The two
connector-specific columns distinguish layer assignment and training
targets.}
\vspace{-2ex}
\label{tab:related_comparison}

\resizebox{0.95\textwidth}{!}{%
\begin{minipage}{\textwidth}

\fontsize{7.5}{8.5}\selectfont

\setlength{\tabcolsep}{2pt}
\renewcommand{\arraystretch}{1.08}

\begin{tabularx}{\linewidth}{@{}
 >{\raggedright\arraybackslash}p{0.145\textwidth}
 >{\centering\arraybackslash}p{0.046\textwidth}
 >{\centering\arraybackslash}p{0.054\textwidth}
 >{\centering\arraybackslash}p{0.049\textwidth}
 >{\raggedright\arraybackslash}p{0.121\textwidth}
 >{\raggedright\arraybackslash}p{0.073\textwidth}
 >{\centering\arraybackslash}p{0.046\textwidth}
 >{\raggedright\arraybackslash}p{0.101\textwidth}
 >{\raggedright\arraybackslash}p{0.134\textwidth}
 >{\raggedright\arraybackslash}X@{}}
\toprule
\textbf{Method} & \textbf{Cross-model phase split} &
\textbf{Inde\-pendent backbones} & \textbf{Hetero. KV geometry} &
\textbf{Cache-layer mapping} & \textbf{Receiver prompt work} &
\textbf{Single decode stream} & \textbf{Adaptation required} &
\textbf{Connector supervision} & \textbf{System evaluation} \\
\midrule

\textbf{\method (ours)} & \(\checkmark\) & \(\checkmark\) &
\(\checkmark\) & Learned local mix / concat & None &
\(\checkmark\) & Stitch; PEFT optional & KV MSE + cosine;
large-anchor KL & Runtime + serving simulation; latency, throughput, energy \\
\midrule

Ridge mapping~\citep{ridge2026} & \(\checkmark\) & \(\checkmark\) &
\(\triangle\) & Fixed top-$k$ ($R^2$) & Skipped &
\(\checkmark\) & Calibration & Receiver KV fit & Runtime:
transfer/prefill timing \\

RaReCache~\citep{rarecache2026} & \(\checkmark\) & \(\checkmark\) &
\(\checkmark\) & Fixed top-$k$ ($R^2$) & Selective / none &
\(\checkmark\) & Per-task calibration & Receiver KV fit &
Serving: queues, dynamic batching, TTFT \\

HeteroFold~\citep{heterofold2026} & \(\checkmark\) & \(\checkmark\) &
\(\checkmark\) & Fixed depth offsets & Skipped &
\(\checkmark\) & Calibration + low-rank corrections & Receiver
attention/output calibration & Runtime: two-GPU handoff \\

Cache-to-Cache~\citep{cachetocache2026} & \(\times\) & \(\checkmark\) &
\(\checkmark\) & Terminal match + gates & Full & \(\checkmark\) &
Fuser & Task-token CE & Runtime: model-pair timing \\

ICaRus~\citep{icarus2026} & \(\times\) & \(\times\) & \(\times\) &
Identity reuse & None & \(\times\) & Decoder PEFT &
--- & Serving: vLLM, QPS sweeps \\

PrefillShare~\citep{prefillshare2026} & \(\checkmark\) & \(\times\) &
\(\times\) & Identity reuse & None & \(\checkmark\) &
Cache-conditioned decoder tuning & --- &
Serving: vLLM, arrival-rate sweeps \\

aLoRA~\citep{alora2025} & \(\times\) & \(\times\) & \(\times\) &
Identity reuse & None & \(\checkmark\) & aLoRA &
--- & Engine: vLLM pipeline timing \\
\midrule

SN-Net~\citep{snnet2023} & --- & \(\checkmark\) & --- & --- & --- & ---
& Stitch layers & --- & Runtime/model: vision throughput/quality \\

ESTA~\citep{esta2024} & --- & \(\checkmark\) & --- & --- & --- & --- &
Stitches + PEFT + biases & --- & Model: vision/FLOPs; training memory
\\

KLAS~\citep{klas2026} & \(\times\) & \(\checkmark\) & --- & --- & Full
hybrid pass & \(\checkmark\) & Stitch selection +
fine-tuning & --- & Model: vision/LLM quality and cost \\

StitchLLM~\citep{stitchllm2025} & \(\times\) & \(\checkmark\) & --- &
--- & Full hybrid pass & \(\checkmark\) & Stitch
layer + router & --- & Serving: cluster/arrival traces \\
\midrule

Standard
speculation~\citep{leviathan2023speculative,chen2023speculative} &
\(\times\) & \(\checkmark\) & --- & --- & Both full & \(\times\) &
None required & --- & Runtime: decoding timing \\

RouteLLM~\citep{routellm2025} & \(\times\) & \(\checkmark\) & --- & ---
& Full & \(\checkmark\) & Router & --- & API cost/quality study \\

FrugalGPT~\citep{frugalgpt2024} & \(\times\) & \(\checkmark\) & --- &
--- & Full per call & \(\triangle\) & Scorer + cascade policy & --- &
API cost/quality study \\
\midrule

MiniLLM~\citep{minillm2024} & \(\times\) &
\(\checkmark\) & --- & --- & Full & \(\checkmark\) &
Student distillation & --- & Model: distillation study \\

SFT~\citep{sft2022} & \(\times\) & --- & --- & --- & Full &
\(\checkmark\) & Full fine-tuning & --- & Model: task quality \\

Adapters~\citep{adapter2019} & --- & --- & --- & --- & --- & --- &
Bottleneck adapters & --- & Model: task/parameter cost \\

LoRA~\citep{lora2022} & \(\times\) & --- & --- & --- & Full &
\(\checkmark\) & Low-rank updates & --- & Runtime: forward timing;
training cost \\

QLoRA~\citep{qlora2023} & \(\times\) & --- & --- & --- & Full &
\(\checkmark\) & Quantized base + LoRA & --- & Model: training/memory
cost \\

DoRA~\citep{dora2024} & \(\times\) & --- & --- & --- & Full &
\(\checkmark\) & Magnitude + low-rank updates & --- & Model:
quality/PEFT cost \\
\midrule

Splitwise~\citep{splitwise2024} & \(\times\) & \(\times\) & \(\times\)
& Identity reuse & None & \(\checkmark\) & None &
--- & Serving: cluster simulation + profiling \\

DistServe~\citep{distserve2024} & \(\times\) & \(\times\) & \(\times\)
& Identity reuse & None & \(\checkmark\) & None &
--- & Serving: cluster goodput/SLOs \\

Prefill-as-a-Service~\citep{prefillasaservice2026} & \(\times\) &
\(\times\) & \(\times\) & Identity reuse & None &
\(\checkmark\) & None & --- & Serving: cross-cluster study/model \\

\bottomrule
\end{tabularx}

\end{minipage}%
}
\end{table*}

\paragraph{Comparison criteria.}
\emph{Method} identifies the cited approach and configuration.
\emph{Cross-model phase split} means that one model's cache replaces a
different decoder's full prompt pass, after which the source is
unnecessary for generation. \emph{Independent backbones} permits
separately pretrained checkpoints without requiring fine-tuned variants
of one base model or a shared cache generator.
\emph{Heterogeneous KV geometry} denotes support for different layer
counts and KV head/feature dimensions. \emph{Cache-layer mapping}
specifies fixed or learned source-layer assignment and aggregation;
identity reuse requires no cache translation.

\emph{Receiver prompt work} counts native processing of shared context
during serving, excluding offline fitting and translation.
\emph{Single decode stream} excludes separate drafting, verification,
or cache-generation streams; a hybrid still qualifies.
\emph{Adaptation required} lists the components fitted or trained,
including any required receiver adapters. \emph{Connector supervision}
identifies the signal used to fit a cache translator or fuser:
receiver KV targets, attention/output targets, native-receiver
predictions, larger-anchor predictions, or task-token labels. It
concerns the cache connector; ordinary model adaptation is listed
separately.

\emph{System evaluation}, the final criterion, describes the scope of
efficiency validation. \emph{Serving} studies include request
arrivals, queues, resource allocation, or shared load, whether
measured or simulated. \emph{Engine} benchmarks execute an
inference-engine pipeline. \emph{Runtime} benchmarks time model,
decoding, or handoff execution without evaluating a serving workload.
\emph{Model} and \emph{API cost/quality} studies report task,
training, memory, or API-cost results. Model-only and isolated runtime
results leave request throughput, queueing delay, and resource
contention under changing load unevaluated. Timing boundaries also
matter: a handoff measured after source prefill is complete does not
include the full cost of a fresh request.

\paragraph{Distinctive features and advantages of \method{}.}
\method{} differs from the closest translators in layer mapping,
training target, and receiver computation.
Within one model family and a shared-tokenizer setting, it connects
separately pretrained checkpoints of different sizes, with different
depths, KV head counts, and head dimensions.
Its weighted-mixture connector learns separate local source-layer
weights for keys and values during training.
Ridge~\citep{ridge2026} and RaReCache~\citep{rarecache2026} instead
select source layers by regression predictability and keep those
selections fixed.
HeteroFold~\citep{heterofold2026} uses fixed relative-depth offsets.
\method{} also provides a concatenation variant and a residual
nonlinear correction, adding flexibility beyond a fixed affine map
while retaining a stable closed-form starting point.

The training target provides a second distinction.
Ridge and RaReCache fit the receiving model's native KV states;
HeteroFold calibrates against its native attention patterns and outputs.
\method{} combines receiver-cache MSE and cosine alignment with KL
supervision from the \emph{larger prefill anchor}.
The cache-alignment loss helps the smaller model use the translated
cache, while the KL loss explicitly trains the connector so that the
smaller model follows the larger model's next-token predictions.
Both anchors remain frozen during this stage.
Receiver PEFT is used only when further task adaptation is needed.
RoPE factoring and closed-form initialization follow prior
work~\citep{ridge2026}; \method{} combines them with learned local
layer mixing, residual nonlinear maps, and larger-anchor supervision.

At inference, \method{} requires one cache handoff and then only the
smaller anchor for generation.
It avoids the full receiver prompt pass used by
Cache-to-Cache~\citep{cachetocache2026}.
RaReCache also reports large-to-small transfer without recomputation,
so this execution pattern alone does not distinguish \method{}.
ICaRus~\citep{icarus2026} and PrefillShare~\citep{prefillshare2026}
train task-specific decoders around a shared base model, while
aLoRA~\citep{alora2025} reuses the cache of its own base model.
\method{} instead translates between different-size pretrained
checkpoints within the same family.
Unlike speculative decoding and block stitching, its generation path
requires neither large-model verification nor large-model blocks.
It supplies the larger anchor's prompt representation for each request,
rather than relying only on offline small-model adaptation or model
selection.

The evaluation includes both real-hardware measurements and a serving
simulation.
On one A40 GPU at batch size 1, a Llama-3.1-Instruct-8B to
Llama-3.2-Instruct-1B request takes 2.96~s, compared with 8.83~s for
the larger anchor alone and 2.80~s for the smaller anchor alone,
with a 1K-token prompt and 256 output tokens.
The serving simulation includes prompt processing, handoff, and
generation, and models arrivals, queues, hardware pools, and transfer.
It extends isolated handoff timings, such as those reported by Ridge
and HeteroFold, to response latency, throughput, and energy estimates
for the complete serving path.
RaReCache also includes serving experiments, with a focus on queued TTFT.
\method{} targets improved quality at a cost close to small-model
serving, without guaranteeing the larger model's output distribution.

\paragraph{Quality gains and further serving efficiency.}
Across six model pairs and five datasets, \method{} improves over
the parameter-matched, fine-tuned small anchor in all 30 settings
and closes up to 80\% of the gap to the larger anchor.
In some cases, its quality is close to that of the larger model:
on CoQA, Qwen2.5-7B to Qwen2.5-3B reaches 81.5 F1, compared with
81.9 for the 7B anchor, while only the 3B model generates the output.
For larger size gaps, the serving simulation reports up to
33.86$\times$ faster response time and 27.5$\times$ lower energy
use for Llama-3.1-Instruct-70B to Llama-3.2-Instruct-1B.
\method{} also fits the prefill/decode worker design of
Splitwise and DistServe~\citep{splitwise2024,distserve2024}.
The simulation already includes phase separation and transfer overlap.
Because the smaller model keeps its native architecture and KV-cache
layout, optimized inference kernels and batching could further improve
efficiency beyond the results reported here.

\section{Limitations and Future work}
\label{sec:limitations}

\method opens a new design point for LLM serving, but several aspects
 remain opportunities for future work. \method currently translates
 the prompt cache once and then lets the decode anchor continue
 natively; richer policies that adapt the handoff during generation
 may further improve quality for long or difficult outputs. Developing
 stitching methods that do not require fine-tuning and provide higher
 accuracy would address a current limitation of the proposed approach.
 Finally, we train one stitching module
 per model pair and dataset, and the two anchors must share a
 tokenizer.

\end{document}